\documentclass[11pt]{article}

\usepackage[final]{acl}
\usepackage{times}
\usepackage{latexsym}

\usepackage[T1]{fontenc}

\usepackage[utf8]{inputenc}

\usepackage{microtype}

\usepackage{inconsolata}

\usepackage{graphicx}
\usepackage{subcaption}
\usepackage{booktabs}
\usepackage{amsfonts}
\usepackage{nicefrac}
\usepackage{amsmath}
\usepackage{amssymb}
\usepackage{mathtools}
\usepackage{algorithm}
\usepackage{algpseudocode}
\usepackage{multirow}
\usepackage{makecell}
\usepackage{bm}
\usepackage{enumitem}
\usepackage{tabularx}
\usepackage{colortbl}
\usepackage{bbm}
\usepackage{pgfplots}
\usepackage{float}
\usepackage{placeins}
\usepackage{caption}

\pgfplotsset{compat=1.18}
\usepackage{xcolor}
\definecolor{myblue}{RGB}{0,102,204}
\definecolor{mygreen}{RGB}{0,140,60}
\definecolor{myred}{RGB}{200,40,40}

\newcommand{\cmark}{\textcolor{mygreen}{$\checkmark$}}

\newcommand{\bppo}{\textsc{BPPO}}
\newcommand{\grpo}{\textsc{GRPO}}
\newcommand{\ppo}{\textsc{PPO}}

\newcommand{\argmin}{\operatorname{argmin}}

\newcommand{\firstn}{\textsc{FirstN}}

\definecolor{darkgreen}{RGB}{0,120,60}
\definecolor{darkred}{RGB}{180,35,35}

\algrenewcommand\algorithmicindent{1.0em}
\definecolor{darkgreen}{RGB}{0,120,60}
\definecolor{darkred}{RGB}{180,35,35}

\usepackage[most]{tcolorbox}

\definecolor{caseblue}{RGB}{36,80,150}
\definecolor{casegray}{RGB}{248,248,248}
\definecolor{grpoRed}{RGB}{210,45,45}
\definecolor{grpoBg}{RGB}{255,238,238}
\definecolor{bppoGreen}{RGB}{45,135,70}
\definecolor{bppoBg}{RGB}{238,250,238}
\definecolor{obsBg}{RGB}{255,250,235}

\definecolor{deltaGreen}{RGB}{0,120,60}
\definecolor{deltaRed}{RGB}{190,40,40}

\definecolor{bestgreen}{RGB}{210,245,215}
\newcommand{\bestbox}[1]{%
  \begingroup
  \setlength{\fboxsep}{2pt}%
  \colorbox{bestgreen}{\strut\textbf{#1}}%
  \endgroup
}

\newtcolorbox{grpowidebox}[1][]{
  enhanced,
  colback=grpoBg,
  colframe=grpoRed,
  coltitle=black,
  colbacktitle=grpoBg,
  fonttitle=\bfseries\small,
  title={\grpo{} Response \cmark},
  boxrule=0.6pt,
  arc=2pt,
  left=6pt,
  right=6pt,
  top=5pt,
  bottom=5pt,
  #1
}

\newtcolorbox{bppowidebox}[1][]{
  enhanced,
  colback=bppoBg,
  colframe=bppoGreen,
  coltitle=black,
  colbacktitle=bppoBg,
  fonttitle=\bfseries\small,
  title={\bppo{} Response \cmark},
  boxrule=0.6pt,
  arc=2pt,
  left=6pt,
  right=6pt,
  top=5pt,
  bottom=5pt,
  #1
}

\newtcolorbox{questionboxwide}[1][]{
  enhanced,
  colback=casegray,
  colframe=caseblue,
  boxrule=0.6pt,
  arc=2pt,
  left=6pt,
  right=6pt,
  top=5pt,
  bottom=5pt,
  #1
}

\newtcolorbox{observationwidebox}[1][]{
  enhanced,
  colback=obsBg,
  colframe=orange!70!black,
  boxrule=0.5pt,
  arc=2pt,
  left=6pt,
  right=6pt,
  top=5pt,
  bottom=5pt,
  #1
}

\title{BPPO: Binary Prefix Policy Optimization for Efficient GRPO-Style Reasoning RL with Concise Responses}

\author{
  Qingfei Zhao\textsuperscript{1,2} \quad
  Huan Song\textsuperscript{1} \quad
  Shuyu Tian\textsuperscript{1} \quad
  Jiawei Shao\textsuperscript{1} \quad
  Xuelong Li\textsuperscript{1}
  \\
  \textsuperscript{1}TeleAI \quad
  \textsuperscript{2}Shanghai Jiao Tong University
}

\begin{document}
\maketitle
\begin{abstract}
Group Relative Policy Optimization (GRPO) is widely used for training reasoning models, but updating all sampled completions in each group incurs substantial cost and can reinforce verbose reasoning trajectories. In this paper, we study whether all completions provide equally useful update signals in GRPO-style reasoning RL. Our gradient-similarity analysis shows that, within the same prompt group, same-class completions often induce highly similar update directions, whereas correct--incorrect pairs provide more distinct contrastive signals. Motivated by this observation, we propose Binary Prefix Policy Optimization (BPPO), which uses the shortest correct completion and the shortest incorrect completion as a compact update unit while preserving full-group advantage normalization. BPPO further improves efficiency with adaptive completion scheduling and prefix-focused optimization; by updating only response prefixes, it avoids reinforcing redundant suffixes and encourages more concise responses. Experiments on GSM8K, MATH, and Geo3K show that BPPO achieves up to \(6.08\times\) speedup over GRPO while maintaining competitive accuracy, and reduces mean response length by approximately 30--50\% without modifying the reward with an explicit length penalty.
\end{abstract}

\section{Introduction}

Large language models (LLMs) have achieved strong performance on multi-step reasoning tasks, especially in mathematics and structured problem solving \citep{wei2022chain,kojima2022large,cobbe2021training,hendrycks2021math}. Reinforcement learning has become an important approach for improving reasoning performance, with policy optimization methods such as \ppo{} and its variants demonstrating substantial empirical gains in instruction following, alignment, and reasoning-oriented optimization \citep{schulman2017proximal,stiennon2020learning,ouyang2022training,bai2022training}. Building on this line of methods, Group Relative Policy Optimization (\grpo{}) avoids training a separate critic model, thereby simplifying the reinforcement learning pipeline and making it a widely adopted approach for reasoning model optimization \citep{shao2024deepseekmath,guo2025deepseek,qwen2025qwq,yu2025dapo,zheng2025gspo}.

\begin{figure}[t]
    \centering
    \includegraphics[width=\columnwidth]{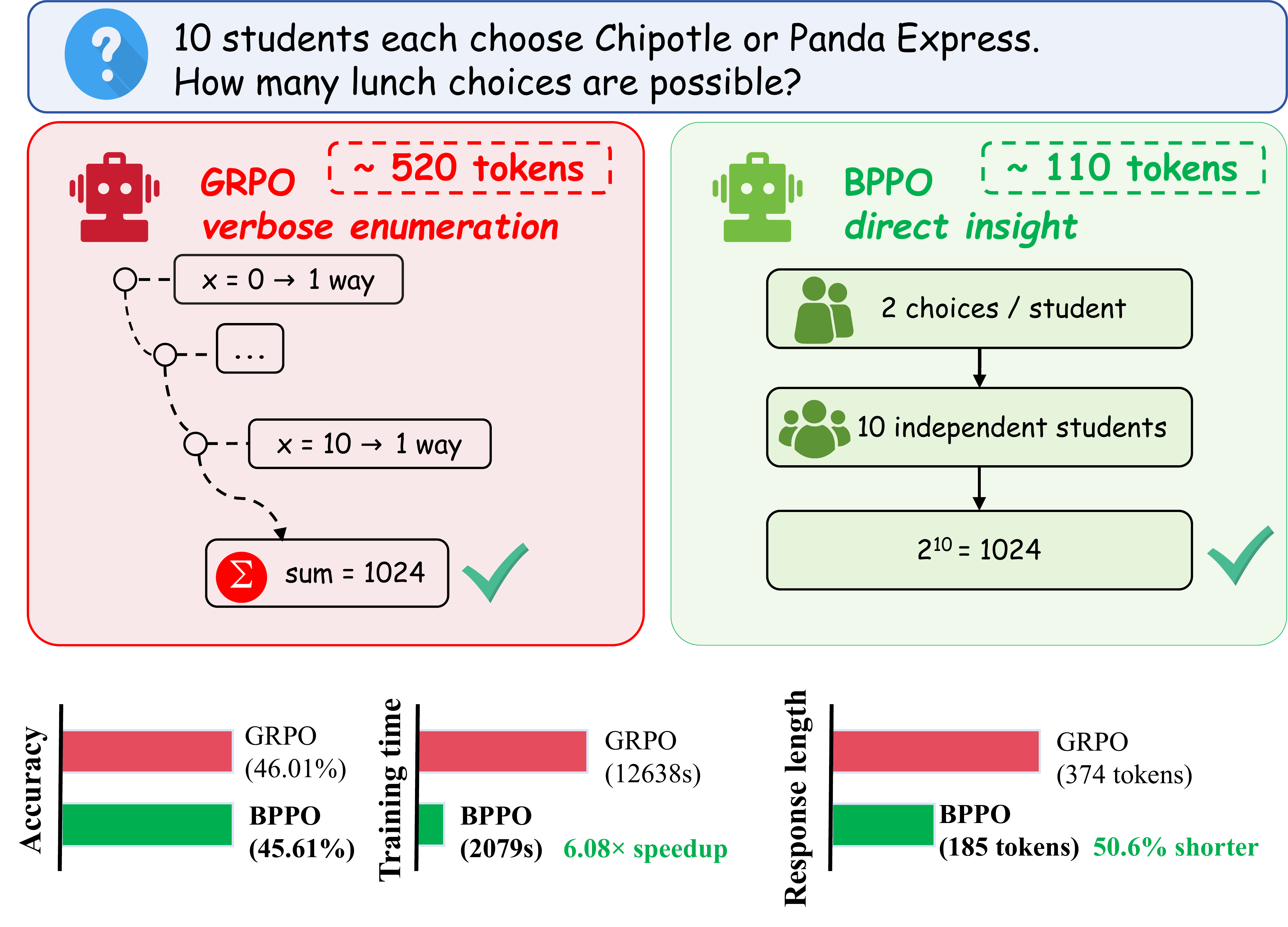}
    \caption{
    A MATH case study from Llama3.2-3B-Instruct trained with \grpo{} and \bppo{}.
    Both methods produce the correct answer, but \grpo{} relies on verbose enumeration, whereas \bppo{} exploits the problem structure and follows a shorter reasoning path.
    In the corresponding MATH setting, \bppo{} maintains comparable accuracy while achieving a 6.08$\times$ training speedup and reducing mean response length by 50.6\%.
    }
    \label{fig:overall_teaser}
\end{figure}

Despite its success, \grpo{} incurs substantial training cost. For each prompt, it samples a group of \(G\) completions, computes rewards for all of them, normalizes advantages within the group, and performs policy updates using the entire group. This design is expensive in two ways. First, the cost grows linearly with the number of sampled completions. Second, reasoning traces are often long, and Transformer attention scales quadratically with sequence length \citep{vaswani2017attention,tay2020efficient}. Beyond training efficiency, full-response updates may also reinforce verbose reasoning patterns~\citep{samineni2025rl, shen-etal-2023-loose, bu-etal-2025-beyond, li-etal-2024-reinforcement, park-etal-2024-disentangling, dumitru-etal-2025-conciserl}. Although efficient attention and memory-management techniques have reduced some practical costs of long-sequence computation \citep{dao2022flashattention,dao2023flashattention2,kwon2023efficient}, updating on all completions and all response tokens in every sampled group remains computationally expensive and may encourage unnecessarily long reasoning trajectories. This motivates more sample- and token-efficient reinforcement learning methods for reasoning models \citep{lin2025cppo,lee2025token,sang2026nat}, and also highlights response length as an important dimension of efficient reasoning optimization.

A natural question is which parts of a sampled group provide the most useful update signals. Our analysis shows that same-prompt completions from the same correctness class often induce highly similar gradients, suggesting overlapping optimization directions. In contrast, correct and incorrect completions yield more distinct update directions, motivating a correct--incorrect pair as a compact update unit. Among such pairs, we select the shortest correct--incorrect pair to preserve cross-class contrast while favoring compact reasoning trajectories. Since early reasoning steps often shape later solution direction, response prefixes can preserve useful token-level update signals without full-response updates. Moreover, avoiding updates on redundant suffixes also reduces reinforcement of unnecessary reflection or repeated verification, which may encourage more concise reasoning trajectories.

Based on these observations, we propose Binary Prefix Policy Optimization (\bppo{}), which keeps full-group reward and advantage computation unchanged but applies policy updates only to the response prefixes of the shortest correct and shortest incorrect completions. This design reduces update-bearing data at both the completion and token levels, while avoiding direct reinforcement of long reasoning trajectories that contain redundant suffixes. To avoid reduced hardware utilization after pair selection, we introduce adaptive completion scheduling, which discards all-correct and all-incorrect groups before importance-ratio computation and packs selected pairs from multiple prompts to maintain efficient device usage.

We evaluate \bppo{} on GSM8K, MATH, and Geo3K using models that span different scales, model families, and modalities, including Qwen2.5-1.5B-Instruct, Llama3.2-3B-Instruct, and Qwen3-VL-4B. Across these settings, \bppo{} achieves up to 5.90$\times$ speedup over \grpo{} on GSM8K, 6.08$\times$ on MATH, and 3.86$\times$ on Geo3K, while maintaining competitive accuracy. In addition, compared with \grpo{}, \bppo{} reduces the mean response length by approximately 30--50\% without modifying the reward with an explicit length penalty.


\section{Related Work}
\label{sec:related}

\paragraph{RL for Reasoning.}
Reinforcement learning has become an important paradigm for improving LLMs. Early successes were largely driven by reinforcement learning with human feedback (RLHF), which substantially improved instruction following and safety \citep{ouyang2022training,bai2022training,stiennon2020learning,rafailov2023direct,meng2024simpo}. More recently, automatically verifiable supervision and reward signals have shown strong potential for enhancing reasoning capabilities, especially in mathematical and other structured domains where correctness can be checked reliably \citep{lightman2023lets,shao2024deepseekmath,guo2025deepseek,qwen2025qwq,wang2024mathshepherd,luo2025reft}. In terms of policy optimization, \ppo{} remains a standard choice \citep{schulman2017proximal}, while \grpo{} has become an important approach for reasoning training because it avoids explicit value modeling through group-relative advantage normalization \citep{shao2024deepseekmath}. Subsequent work further demonstrated that \grpo{}-style training can induce strong reasoning behaviors in LLMs \citep{guo2025deepseek,qwen2025qwq}, and also inspired a number of follow-up variants and refinements. For example, DAPO improves scalability and training stability for large-scale reasoning RL \citep{yu2025dapo}, while GSPO revisits the optimization objective from a sequence-level perspective and reports improved efficiency and stability over \grpo{} \citep{zheng2025gspo}.

\paragraph{Acceleration Methods for Reasoning RL.}
The high cost of reasoning RL has motivated recent efforts to improve the efficiency and scalability of \grpo{}-style training \citep{zheng2026act, NEURIPS2025_fd54731c,fu2026areal,chen2026the}. DAPO improves reasoning RL through several system and algorithmic refinements, including dynamic sampling and overlong filtering \citep{yu2025dapo}. Other work reduces unnecessary computation more directly. CPPO accelerates \grpo{} by pruning completions with low contribution to policy updates, reducing the number of trajectories used in optimization \citep{lin2025cppo}. Token-efficient RL methods further update only selected subsets of response tokens to reduce computation cost \citep{lee2025token,sang2026nat,mundada2026wsgrpo}. These studies suggest that efficiency in reasoning RL can be improved not only through system-level optimization, but also by selectively reducing completion- or token-level updates while preserving useful learning signals, highlighting the need to identify which trajectories and tokens are most informative for policy optimization.

\paragraph{Response-Length Control.}
Long reasoning traces can improve multi-step problem solving, but they also increase training and inference cost and may contain redundant derivations or repeated verification \citep{liu2026learn,fang2026thinkless,dai2026sgrpo}. Prior work has observed that RLHF and RL post-training can favor longer or more verbose responses, partly because sequence-level rewards provide coarse feedback, resulting in coarse credit assignment that may reinforce all sampled tokens in positively rewarded completions \citep{singhal2023long,samineni2025rl,liu2025understanding}. To control response length, recent methods introduce explicit length-aware mechanisms, including overlong filtering \citep{yu2025dapo}, adaptive length penalties and reward shaping \citep{li2025leash,xiang2025just}, length-constrained RL objectives \citep{aggarwal2025l1}, RL-based chain-of-thought pruning \citep{hou2025thinkprune}, and dynamic allocation of inference compute \citep{arora2025training}. These works highlight response length as an important dimension of reasoning-RL optimization, alongside accuracy and training efficiency.

\section{Method}
\label{sec:method}

\subsection{Group Relative Policy Optimization}
\label{sec:grpo}

\grpo{} simplifies reinforcement learning by removing the need for a separate critic and computing relative advantages from rewards within each sampled completion group. For each prompt \(q \sim \mathcal{D}\), it samples \(G\) completions \(\{o_1,\ldots,o_G\} \sim \pi_{\theta_{\mathrm{old}}}(\cdot \mid q)\), computes rewards for all completions, and normalizes them within the group to obtain completion-level advantages. The advantage \(\hat{A}_i\) is then applied to completion \(o_i\) during token-level policy optimization. The token-level importance ratio is
\(\rho_{i,t}(\theta)=\pi_\theta(o_{i,t}\mid q,o_{i,<t})/\pi_{\theta_{\mathrm{old}}}(o_{i,t}\mid q,o_{i,<t})\),
where \(o_{i,<t}\) denotes the response prefix before token \(t\). The current policy \(\pi_\theta\) is optimized against a frozen reference policy \(\pi_{\mathrm{ref}}\) with a clipped policy-gradient objective and an additional KL regularization term:

\begingroup
\small
\begin{equation}
\label{eq:grpo_obj}
\begin{aligned}
J(\theta)
=&\;
\mathbb{E}_{q\sim\mathcal{D},\,\{o_i\}_{i=1}^{G}\sim\pi_{\theta_{\mathrm{old}}}(\cdot\mid q)}
\Bigg[
\frac{1}{G}\sum_{i=1}^{G}
\Bigg(
\frac{1}{|o_i|}\sum_{t=1}^{|o_i|}
\Bigl[
\\
&\;
\min\Big(
\rho_{i,t}(\theta)\hat{A}_i,\,
\mathrm{clip}(\rho_{i,t}(\theta),1-\varepsilon,1+\varepsilon)\hat{A}_i
\Big)
\\
&\;
-\beta\,\mathbb{D}_{\mathrm{KL}}
\!\left[\pi_\theta \,\|\, \pi_{\mathrm{ref}}\right]
\Bigr]
\Bigg)
\Bigg].
\end{aligned}
\end{equation}
\endgroup

where $\varepsilon$ and $\beta$ are hyperparameters, the KL term is defined as
\begin{equation}
\begin{aligned}
\mathbb{D}_{\mathrm{KL}}(\pi_\theta \| \pi_{\mathrm{ref}})
=&\;
\frac{\pi_{\mathrm{ref}}(o_{i,t}\mid q,o_{i,<t})}
{\pi_\theta(o_{i,t}\mid q,o_{i,<t})}
\\
&\;
-\log
\frac{\pi_{\mathrm{ref}}(o_{i,t}\mid q,o_{i,<t})}
{\pi_\theta(o_{i,t}\mid q,o_{i,<t})}
-1 .
\end{aligned}
\end{equation}
and the advantage $\hat{A}_i$ is computed using a group of rewards $\{r_1, r_2, \ldots, r_G\}$:
\begin{equation}
\hat{A}_i =
\frac{r_i - \mathrm{mean}(\{r_1, r_2, \ldots, r_G\})}
{\mathrm{std}(\{r_1, r_2, \ldots, r_G\})}.
\end{equation}

\subsection{Gradient Redundancy}
\label{sec:grad_redundancy}
To better understand how useful update signals are distributed within a sampled group, we analyze gradient similarity among completions sampled for the same prompt in \grpo{} training. Using Qwen2.5-1.5B-Instruct on GSM8K, we conduct this analysis under three sampling temperatures: \(T=0.8\), \(0.9\), and \(1.0\). At each temperature, we sample \(G=16\) completions per prompt.

\begin{figure*}[!t]
    \centering

    \begin{subfigure}[b]{0.32\textwidth}
        \centering
        \includegraphics[width=0.88\linewidth]{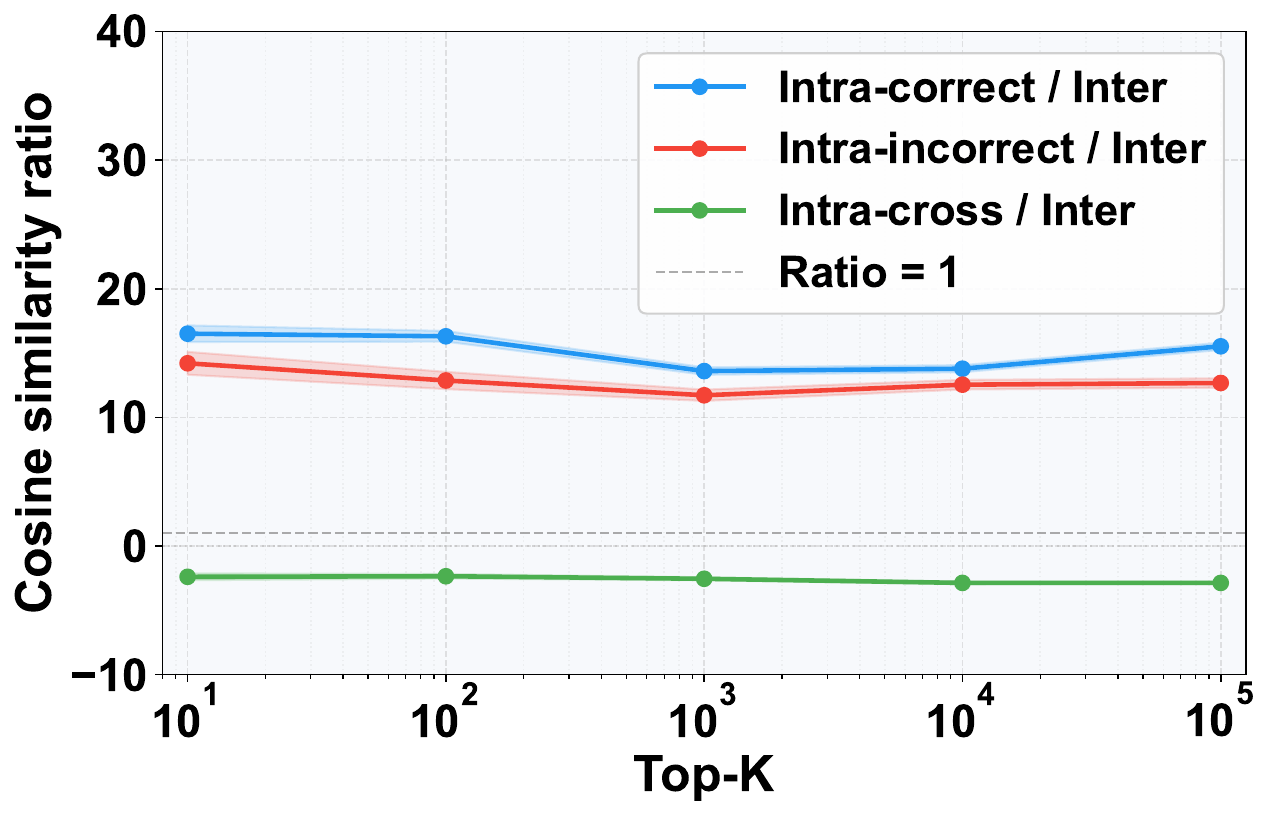}
        \caption{Cosine ratio, \(T=0.8\)}
    \end{subfigure}
    \hfill
    \begin{subfigure}[b]{0.32\textwidth}
        \centering
        \includegraphics[width=0.88\linewidth]{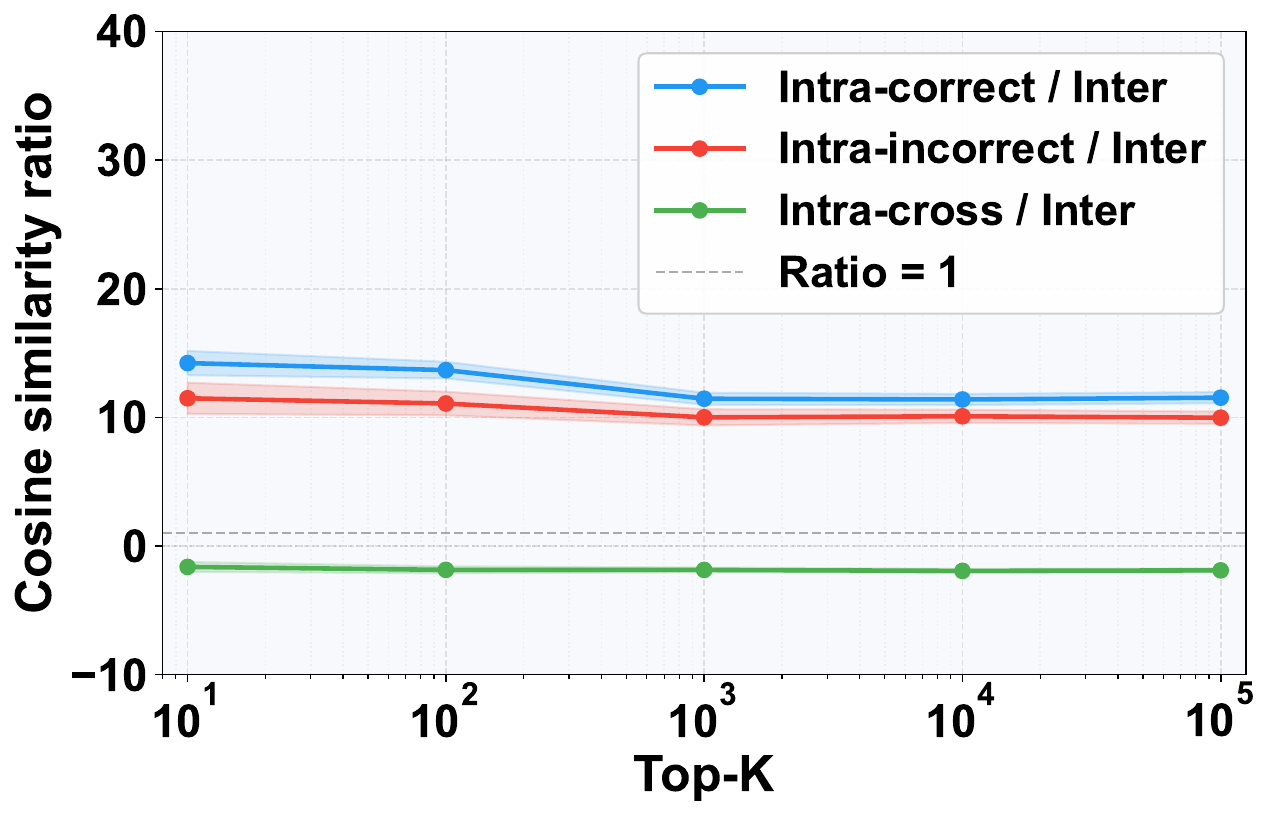}
        \caption{Cosine ratio, \(T=0.9\)}
    \end{subfigure}
    \hfill
    \begin{subfigure}[b]{0.32\textwidth}
        \centering
        \includegraphics[width=0.88\linewidth]{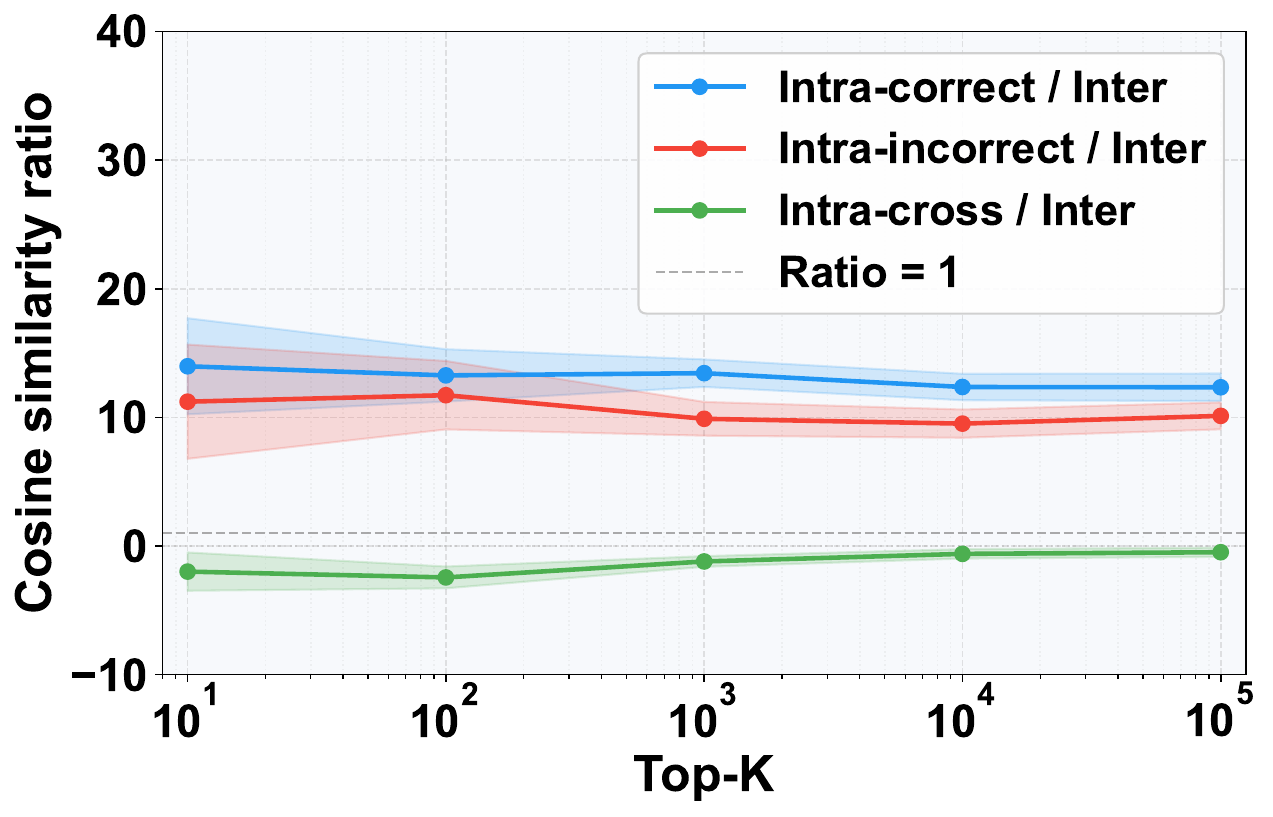}
        \caption{Cosine ratio, \(T=1.0\)}
    \end{subfigure}

    \vspace{0.4em}

    \begin{subfigure}[b]{0.32\textwidth}
        \centering
        \includegraphics[width=0.8\linewidth]{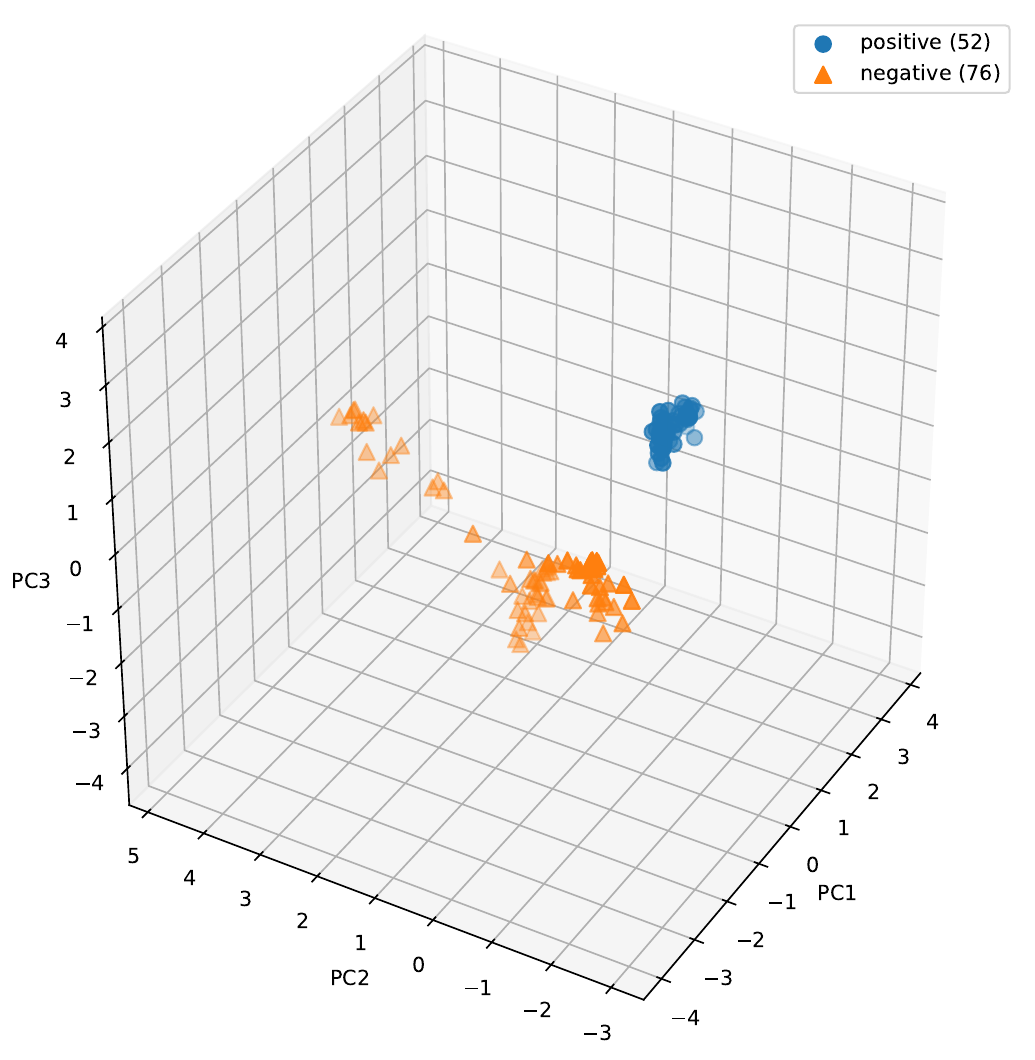}
        \caption{PCA, \(T=0.8\)}
    \end{subfigure}
    \hfill
    \begin{subfigure}[b]{0.32\textwidth}
        \centering
        \includegraphics[width=0.8\linewidth]{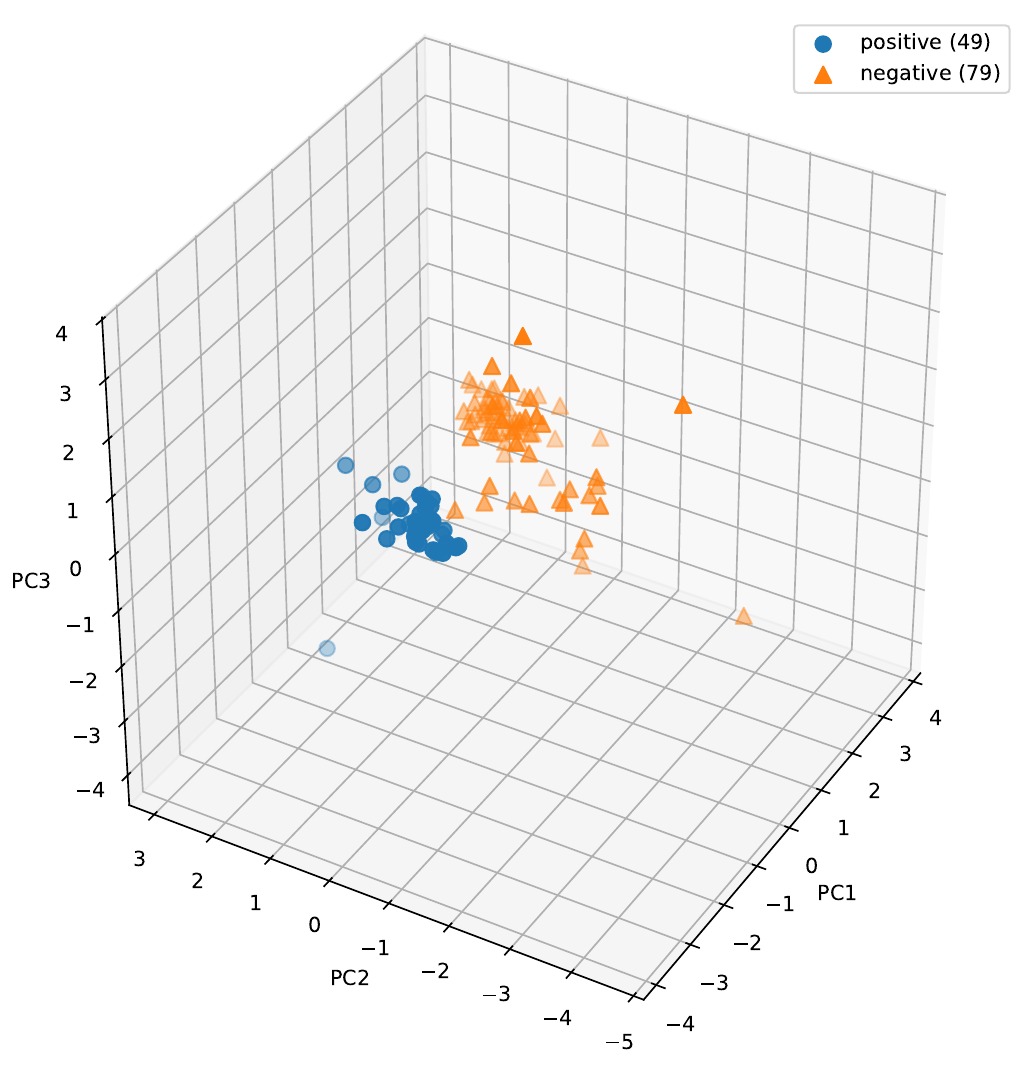}
        \caption{PCA, \(T=0.9\)}
    \end{subfigure}
    \hfill
    \begin{subfigure}[b]{0.32\textwidth}
        \centering
        \includegraphics[width=0.8\linewidth]{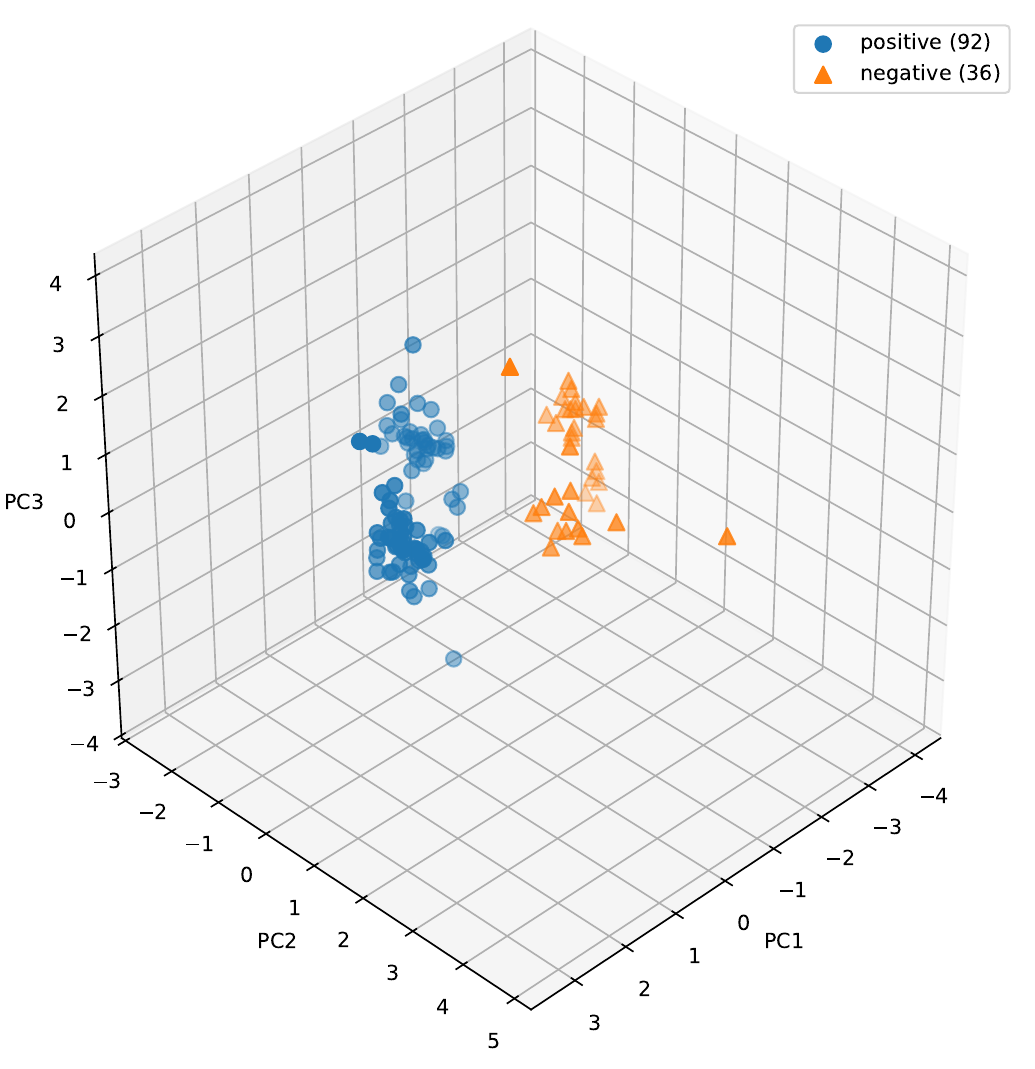}
        \caption{PCA, \(T=1.0\)}
    \end{subfigure}
    \caption{
    \textbf{Gradient similarity under different sampling temperatures.}
    (a)--(c) show top-\(K\) gradient cosine-similarity ratios for \(T=0.8\), \(0.9\), and \(1.0\), with shaded regions indicating \(3\sigma\) intervals. Each ratio normalizes same-prompt gradient similarity by the average similarity between completions from different prompts.
    (d)--(f) show PCA projections of gradient vectors from 128 completions sampled from same prompt.
    Across temperatures, same-class pairs exhibit substantially higher similarity, whereas correct--incorrect pairs remain more distinct.
    }
    \label{fig:gradient_structure}
\end{figure*}

\paragraph{Top-\(K\) Cosine Similarity Ratios.}
For each completion, we compute its gradient, retain the top-\(K\) entries with the largest absolute values for \(K\in\{10^1,10^2,10^3,10^4,10^5\}\), and measure cosine similarity on the truncated vector. We consider three types of same-prompt completion pairs: correct--correct, incorrect--incorrect, and correct--incorrect. 
For each type, we report the ratio between its cosine similarity and the average similarity of completion pairs from different prompts.  
Figure~\ref{fig:gradient_structure} reports these ratios as Intra-correct / Inter, Intra-incorrect / Inter, and Intra-cross / Inter. Correct--correct and incorrect--incorrect pairs show much higher similarity ratios across temperatures, while correct--incorrect pairs remain low and negative. The PCA projections further reveal a class-dependent gradient structure, suggesting that same-class completions provide overlapping update directions, whereas correct--incorrect pairs retain distinctive contrastive signals.




\subsection{Pair Selection}
\label{sec:pair_selection}

The gradient-similarity analysis suggests that same-class completions often provide overlapping update directions, while correct--incorrect pairs provide more distinctive contrastive signals. Therefore, \bppo{} restricts policy updates to one selected correct completion and one selected incorrect completion from each sampled group.

Let \(\mathcal{I}^+\) and \(\mathcal{I}^-\) denote the sets of correct and incorrect completions, respectively, and let \(\ell(o_i)\) be the response length of completion \(o_i\). \bppo{} selects one completion from each class to form the update set \(\mathcal{S}(q)=\{i^*_+, i^*_-\}\). 
Specifically, it selects the shortest completion from each correctness class, preserving correctness contrast while favoring compact update trajectories. As shown in Section~\ref{sec:ablation}, this strategy achieves a favorable trade-off among accuracy, training efficiency, and response conciseness:
\begin{equation}
\label{eq:pair_selection}
i^*_+ = \arg\min_{i \in \mathcal{I}^+} \ell(o_i),
\quad
i^*_- = \arg\min_{i \in \mathcal{I}^-} \ell(o_i).
\end{equation}

\subsection{\firstn{}: Prefix-Focused Gradient Update}
\label{sec:firstn}

While pair selection reduces the number of update-bearing completions, long reasoning traces can still make token-level optimization costly. Since early tokens often guide reasoning direction, we introduce \firstn{}, which restricts policy-gradient updates to the first \(n\) response tokens of each selected completion. Full-response generation and reward computation remain unchanged, while updates are applied only to response prefixes. This preserves full-rollout reward supervision, reduces optimization cost on long responses, and avoids reinforcing redundant suffixes such as unnecessary reflection or repeated verification, which may encourage more concise responses. We study the effect of prefix length in Section~\ref{sec:ablation}.

\begin{figure*}[t]
    \centering
    \includegraphics[width=0.95\textwidth]{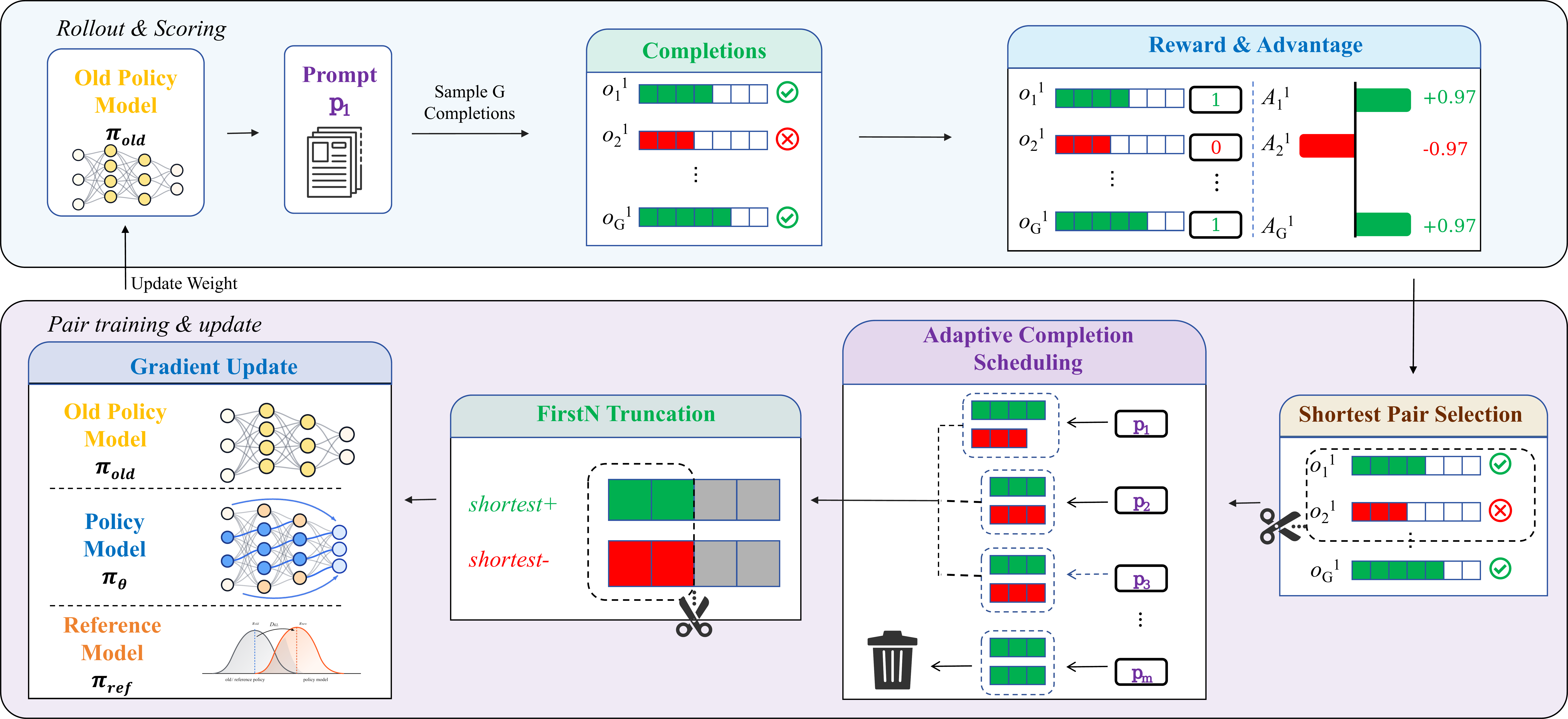}
    \caption{
    \textbf{Overview of the \bppo{} training workflow.}
    The top panel shows rollout and scoring with full-group rewards and advantages as in \grpo{}.
    The bottom panel shows pair-based training: each prompt first retains the shortest correct--incorrect pair, discards groups without both correctness classes, packs retained pairs from multiple prompts into the update batch, and applies prefix-focused updates to the retained completions.
    }
    \label{fig:adaptive_completion_scheduling}
\end{figure*}

\subsection{BPPO}
\label{sec:bppo_overall}

Binary Prefix Policy Optimization (\bppo{}) combines selected-pair updates at the completion level with prefix-focused updates at the token level. Specifically, \bppo{} restricts policy updates to the shortest correct completion and the shortest incorrect completion from each sampled group, using them as a compact contrastive update unit, and further applies policy-gradient updates only to their response prefixes. Relative to \grpo{}, \bppo{} preserves the same sampling process, reward computation, and group-relative advantage normalization, changing only where policy updates are applied. Let \(\mathcal{S}(q)=\{i^*_+,i^*_-\}\) denote the selected shortest correct--incorrect pair for prompt \(q\), and let \(n\) denote the number of response tokens used for prefix-focused optimization. The resulting \bppo{} objective is:

\begingroup
\small
\begin{equation}
\label{eq:bppo_obj}
\begin{aligned}
J(\theta)
=&\;
\mathbb{E}_{q\sim\mathcal{D},\,\{o_i\}_{i=1}^{G}\sim\pi_{\theta_{\mathrm{old}}}(\cdot\mid q)}
\Bigg[
\frac{1}{|\mathcal{S}(q)|}
\sum_{i\in\mathcal{S}(q)}
\Bigg(
\frac{1}{n}\sum_{t=1}^{n}
\Bigl[
\\
&\;
\min\Big(
\rho_{i,t}(\theta)\hat{A}_i,\,
\mathrm{clip}(\rho_{i,t}(\theta),1-\varepsilon,1+\varepsilon)\hat{A}_i
\Big)
\\
&\;
-\beta\,\mathbb{D}_{\mathrm{KL}}
\!\left[\pi_\theta \,\|\, \pi_{\mathrm{ref}}\right]
\Bigr]
\Bigg)
\Bigg].
\end{aligned}
\end{equation}
\endgroup

\paragraph{Adaptive Completion Scheduling.}
Pair selection leaves each prompt with at most one retained correct--incorrect pair, reducing update-bearing completions under fixed prompt batches. To maintain GPU utilization, adaptive completion scheduling aggregates retained pairs from additional prompts until the per-device update budget matches the original budget, while discarding all-correct and all-incorrect groups before importance-ratio computation. This improves utilization and reduces optimization steps per epoch without changing the pair-selection rule or prefix-update objective. The procedure is shown in Figure~\ref{fig:adaptive_completion_scheduling} and Appendix~\ref{app:algorithm}; Appendix~\ref{app:acs_effect} reports matched-step results without adaptive scheduling.

\section{Experiments}
\label{sec:experiments}

\subsection{Experimental Settings}
\label{sec:setup}

\begin{table*}[t]
  \centering
  \small
  \setlength{\tabcolsep}{4.5pt}
  \renewcommand{\arraystretch}{1.06}
    \caption{
    \textbf{Results on GSM8K with Qwen2.5-1.5B-Instruct under different rollout group sizes \(G\)}.
    \bppo{} achieves up to \(5.90\times\) speedup and reduces mean response length by up to \(43.9\%\) while maintaining competitive accuracy. Negative length reduction indicates longer responses than \grpo{}.
    }
  \label{tab:gsm8k_main}
  \begin{tabular*}{0.98\textwidth}{@{\extracolsep{\fill}}lcccccc}
    \toprule
    \textbf{Method} 
    & \textbf{\makecell{Group\\Size \(G\)}} 
    & \textbf{Accuracy (\%)} 
    & \textbf{\makecell{Mean\\Response Tokens}} 
    & \textbf{\makecell{Length\\Reduction (\%)}} 
    & \textbf{\makecell{Training\\Time (s)}} 
    & \textbf{\makecell{Speedup\\(times)}} \\
    \midrule
    Qwen2.5-1.5B-Instruct & -  & 23.63 & - & - & - & - \\
    \midrule

    \grpo{} & \multirow{5}{*}{8}  
    & \(75.65 \pm 0.43\) 
    & \(259.05 \pm 1.72\) 
    & -
    & \(7186 \pm 1339\) 
    & 1.00 \\

    CPPO
    &
    & \(75.50 \pm 0.40\)
    & \(285.72 \pm 6.84\)
    & -10.3
    & \(2361 \pm 93\)
    & \(3.04 \pm 0.12\) \\

    \grpo{}+\firstn{} 
    &                       
    & \(75.84 \pm 0.54\)
    & \(232.01 \pm 7.60\)
    & 10.4
    & \(6002 \pm 531\) 
    & \(1.20 \pm 0.17\) \\

    Pair 
    &                       
    & \(76.24 \pm 0.16\)
    & \(177.21 \pm 2.06\)
    & 31.6
    & \(2008 \pm 322\) 
    & \(3.58 \pm 0.57\) \\

    \bppo{} 
    &                       
    & \(75.34 \pm 0.37\)
    & \(155.06 \pm 3.32\)
    & \bestbox{40.1}
    & \(1812 \pm 20\) 
    & \bestbox{\(\mathbf{3.97} \pm 0.04\)} \\

    \midrule

    \grpo{} & \multirow{5}{*}{16} 
    & \(76.57 \pm 0.57\) 
    & \(262.64 \pm 2.38\) 
    & -
    & \(11816 \pm 515\) 
    & 1.00 \\

    CPPO
    &
    & \(76.85 \pm 0.59\)
    & \(265.70 \pm 6.68\)
    & -1.2
    & \(2541 \pm 312\)
    & \(4.65 \pm 0.54\) \\

    \grpo{}+\firstn{} 
    &                       
    & \(76.55 \pm 0.50\)
    & \(253.17 \pm 7.10\)
    & 3.6
    & \(11197 \pm 174\) 
    & \(1.06 \pm 0.02\) \\

    Pair 
    &                       
    & \(76.48 \pm 0.42\)
    & \(161.03 \pm 1.77\)
    & 38.7
    & \(2271 \pm 18\) 
    & \(5.20 \pm 0.04\) \\

    \bppo{} 
    &                       
    & \(75.92 \pm 0.38\)
    & \(147.25 \pm 4.21\)
    & \bestbox{43.9}
    & \(2002 \pm 49\) 
    & \bestbox{\(\mathbf{5.90} \pm 0.24\)} \\

    \bottomrule
  \end{tabular*}
\end{table*}

\begin{table*}[t]
  \centering
  \small
  \setlength{\tabcolsep}{4.5pt}
  \renewcommand{\arraystretch}{1.06}
    \caption{
    \textbf{Results on MATH with Llama3.2-3B-Instruct under different rollout group sizes \(G\).}
    \bppo{} achieves up to \(6.08\times\) speedup and reduces mean response length by up to \(50.6\%\) while maintaining competitive accuracy.
    }
  \label{tab:math_main}
  \begin{tabular*}{0.98\textwidth}{@{\extracolsep{\fill}}lcccccc}
    \toprule
    \textbf{Method} 
    & \textbf{\makecell{Group\\Size \(G\)}} 
    & \textbf{Accuracy (\%)} 
    & \textbf{\makecell{Mean\\Response Tokens}} 
    & \textbf{\makecell{Length\\Reduction (\%)}} 
    & \textbf{\makecell{Training\\Time (s)}} 
    & \textbf{\makecell{Speedup\\(times)}} \\
    \midrule
    Llama3.2-3B-Instruct & -  & 33.12 & - & - & - & - \\
    \midrule

    \grpo{} & \multirow{5}{*}{8}  
    & \(45.33 \pm 0.49\) 
    & \(360.54 \pm 3.01\) 
    & -
    & \(9070 \pm 244\)  
    & 1.00 \\

    CPPO
    &
    & \(45.88 \pm 0.35\)
    & \(417.70 \pm 3.12\)
    & -15.9
    & \(3502 \pm 63\)
    & \(2.59 \pm 0.05\) \\

    \grpo{}+\firstn{} 
    &                       
    & \(44.91 \pm 0.20\)
    & \(319.77 \pm 37.44\)
    & 11.3
    & \(7904 \pm 596\)  
    & \(1.15 \pm 0.06\) \\

    Pair 
    &                       
    & \(45.68 \pm 0.32\)
    & \(233.00 \pm 8.73\)
    & 35.4
    & \(3207 \pm 42\)  
    & \(2.83 \pm 0.04\) \\

    \bppo{} 
    &                       
    & \(45.23 \pm 0.14\)
    & \(190.00 \pm 7.50\)
    & \bestbox{47.3}
    & \(2788 \pm 22\)  
    & \bestbox{\(\mathbf{3.25} \pm 0.03\)} \\

    \midrule

    \grpo{} & \multirow{5}{*}{16} 
    & \(46.01 \pm 0.77\) 
    & \(374.83 \pm 17.24\) 
    & -
    & \(12638 \pm 878\) 
    & 1.00 \\

    CPPO
    &
    & \(46.19 \pm 0.50\)
    & \(414.48 \pm 11.60\)
    & -10.6
    & \(2408 \pm 179\)
    & \(5.25 \pm 0.38\) \\

    \grpo{}+\firstn{} 
    &                       
    & \(46.38 \pm 0.50\)
    & \(362.06 \pm 26.02\)
    & 3.4
    & \(11775 \pm 813\) 
    & \(1.07 \pm 0.14\) \\

    Pair 
    &                       
    & \(46.14 \pm 0.21\)
    & \(192.74 \pm 14.06\)
    & 48.6
    & \(2249 \pm 46\) 
    & \(5.62 \pm 0.41\) \\

    \bppo{} 
    &                       
    & \(45.61 \pm 0.15\)
    & \(185.20 \pm 2.95\)
    & \bestbox{50.6}
    & \(2079 \pm 73\) 
    & \bestbox{\(\mathbf{6.08} \pm 0.55\)} \\

    \bottomrule
  \end{tabular*}
\end{table*}

\paragraph{Training Details.}
We implement \bppo{} within the verl framework \citep{sheng2024hybridflow} and use vLLM \citep{kwon2023efficient} for efficient completion generation. We experiment with Qwen2.5-1.5B-Instruct \citep{qwen2024qwen25}, Llama3.2-3B-Instruct \citep{grattafiori2024llama3}, and Qwen3-VL-4B \citep{bai2025qwen3vl}, using one H100 GPU for the 1.5B model and four H100 GPUs for larger models. Unless otherwise specified, all models are trained for one epoch with sampling temperature 1.0 and learning rate \(1\times10^{-6}\). We use rollout group sizes \(G\in\{8,16,32\}\), report the \(G=8\) and \(G=16\) results in the main text, and provide \(G=32\) results in Appendix~\ref{app:g32_results}. We set \(G=16\) by default for ablations. For CPPO \citep{lin2025cppo}, we use pruning rates of 75\% for \(G=8\) and 87.5\% for \(G=16\) to match \bppo{}'s retained-completion budget. Each setting is trained three times independently, and we report mean \(\pm\) standard deviation. The maximum completion length is set to 2048 tokens for GSM8K and MATH, and to 8192 tokens for Geo3K. For all experiments involving \firstn{}, we update only the first \(n\) response tokens, where \(n\) is set to 50\% of the average response length.

\paragraph{Evaluation Details.}
We evaluate the main results on three reasoning benchmarks: MATH \citep{hendrycks2021math}, GSM8K \citep{cobbe2021training}, and Geo3K \citep{lu2021intergps}. We further report additional evaluations on AIME24 \citep{maa2024aime}, MinervaMATH \citep{lewkowycz2022solving}, and OlympiadBench \citep{he2024olympiadbench} in Appendix~\ref{app:additional_math_benchmarks}. Performance is measured by Pass@1 accuracy under greedy decoding.

\subsection{Main Results}
\label{sec:main_results}

We evaluate \bppo{} on GSM8K, MATH, and Geo3K. We compare \textbf{\grpo{}}, \textbf{CPPO} \citep{lin2025cppo}, and three controlled variants: \textbf{\grpo{}+\firstn{}}, which applies prefix-focused updates to all completions; \textbf{Pair}, which updates only on the shortest correct--incorrect pair; and \textbf{\bppo{}}, which combines pair selection with prefix-focused optimization. Further larger-model results and case studies are provided in Appendices~\ref{app:larger_model} and~\ref{app:case_study}.


\begin{table*}[t]
  \centering
  \small
  \setlength{\tabcolsep}{4.5pt}
  \renewcommand{\arraystretch}{1.06}
    \caption{
    \textbf{Results on Geo3K with Qwen3-VL-4B under different rollout group sizes \(G\).}
    \bppo{} achieves up to \(3.86\times\) speedup and reduces mean response length by up to \(45.7\%\) while maintaining competitive accuracy.
    }
  \label{tab:geo3k_main}
  \begin{tabular*}{0.98\textwidth}{@{\extracolsep{\fill}}lcccccc}
    \toprule
    \textbf{Method} 
    & \textbf{\makecell{Group\\Size \(G\)}} 
    & \textbf{Accuracy (\%)} 
    & \textbf{\makecell{Mean\\Response Tokens}} 
    & \textbf{\makecell{Length\\Reduction (\%)}} 
    & \textbf{\makecell{Training\\Time (s)}} 
    & \textbf{\makecell{Speedup\\(times)}} \\
    \midrule
    Qwen3-VL-4B & -  & 31.82 & - & - & - & - \\
    \midrule

    \grpo{} & \multirow{5}{*}{8} 
    & \(48.12 \pm 0.82\) 
    & \(3049.95 \pm 19.66\) 
    & -
    & \(25071 \pm 166\) 
    & 1.00 \\

    CPPO
    &
    & \(48.22 \pm 0.90\)
    & \(4073.86 \pm 106.60\)
    & -33.6
    & \(7750 \pm 142\)
    & \(3.23 \pm 0.07\) \\

    \grpo{}+\firstn{} 
    &                       
    & \(47.12 \pm 0.44\)
    & \(3016.77 \pm 104.00\)
    & 1.1
    & \(22981 \pm 39\) 
    & \(1.09 \pm 0.01\) \\

    Pair              
    &                       
    & \(49.67 \pm 0.52\)
    & \(2136.67 \pm 19.65\)
    & 29.9
    & \(6912 \pm 94\) 
    & \(3.63 \pm 0.04\) \\

    \bppo{} 
    &                       
    & \(49.43 \pm 0.50\)
    & \(2063.23 \pm 22.77\)
    & \bestbox{32.4}
    & \(6490 \pm 172\)  
    & \bestbox{\(\mathbf{3.86} \pm 0.10\)} \\

    \midrule

    \grpo{} & \multirow{5}{*}{16} 
    & \(48.87 \pm 0.23\) 
    & \(3066.45 \pm 27.28\) 
    & -
    & \(29264 \pm 184\) 
    & 1.00 \\

    CPPO
    &
    & \(49.43 \pm 0.85\)
    & \(3851.84 \pm 63.57\)
    & -25.6
    & \(11773 \pm 125\)
    & \(2.49 \pm 0.03\) \\

    \grpo{}+\firstn{} 
    &                       
    & \(47.97 \pm 0.48\)
    & \(3006.12 \pm 54.54\)
    & 2.0
    & \(27007 \pm 256\)  
    & \(1.08 \pm 0.01\) \\

    Pair 
    &                       
    & \(49.75 \pm 0.36\)
    & \(1830.37 \pm 21.08\)
    & 40.3
    & \(11388 \pm 253\) 
    & \(2.57 \pm 0.04\) \\

    \bppo{} 
    &                       
    & \(48.93 \pm 0.48\)
    & \(1665.13 \pm 7.38\)
    & \bestbox{45.7}
    & \(10710 \pm 108\) 
    & \bestbox{\(\mathbf{2.73} \pm 0.01\)} \\

    \bottomrule
  \end{tabular*}
\end{table*}

\paragraph{Training on GSM8K.}

On GSM8K, Pair substantially reduces training time under both group sizes while maintaining competitive accuracy. Building on Pair, \bppo{} further applies prefix-focused optimization and achieves up to \(5.90\times\) speedup at \(G=16\), with only minor accuracy changes relative to \grpo{}. Compared with CPPO, \bppo{} achieves higher speedups and much shorter responses across group sizes. At \(G=16\), Pair already reduces mean response length by \(38.7\%\) relative to \grpo{}, while adding \firstn{} further reduces it from \(161.03\) to \(147.25\) tokens, yielding a total reduction of \(43.9\%\). Prefix-focused optimization further shortens responses by concentrating updates on early reasoning decisions and avoiding direct reinforcement of redundant suffixes, such as unnecessary reflection or repeated verification. Overall, these results suggest that updating only the response prefixes of the shortest pair can achieve competitive accuracy with \grpo{}, while improving training efficiency and producing more concise responses.

\paragraph{Training on MATH.}

As shown in Table~\ref{tab:math_main}, \bppo{} remains effective on MATH, a more challenging reasoning benchmark, achieving up to \(6.08\times\) speedup over \grpo{} while maintaining competitive accuracy. Compared with CPPO, \bppo{} achieves higher speedups and much more concise responses across both rollout group sizes; at \(G=16\), \bppo{} improves the speedup from \(5.25\times\) to \(6.08\times\) and reduces the mean response length by \(50.6\%\), whereas CPPO increases response length by \(10.6\%\). These results show that \bppo{} provides a favorable efficiency--conciseness trade-off even on a more challenging mathematical reasoning benchmark.

\paragraph{Training on Geo3K.}

We further evaluate \bppo{} on the multimodal Geo3K benchmark using Qwen3-VL-4B. As shown in Table~\ref{tab:geo3k_main}, \bppo{} remains effective in this setting, achieving up to \(3.86\times\) speedup over \grpo{} while maintaining competitive accuracy and producing substantially more concise responses. Compared with CPPO, \bppo{} achieves higher speedups and much shorter responses under both rollout group sizes; at \(G=16\),  \bppo{} improves the speedup from \(2.49\times\) to \(2.73\times\) and reduces the mean response length by \(45.7\%\) relative to \grpo{}, whereas CPPO increases the mean response length by \(25.6\%\). These results show that \bppo{} provides consistent efficiency and conciseness gains in the multimodal reasoning setting.

\subsection{Compatibility with Other RL Objectives}
\label{sec:compatibility}

Since \bppo{} only restricts update-bearing completions and tokens, it can be naturally combined with other reasoning-RL methods. 
Applying it to DAPO and GSPO preserves their original reward and advantage designs while substantially reducing training time with only minor accuracy changes, as shown in Table~\ref{tab:compatibility_main}. 
Full results are provided in Appendix~\ref{app:compatibility}.

\begin{table}[H]
  \centering
  \footnotesize
  \setlength{\tabcolsep}{4pt}
  \renewcommand{\arraystretch}{1.08}
    \caption{
    Compatibility results of \bppo{} with DAPO on MATH using Qwen3-8B and with GSPO on GSM8K using DeepSeek-R1-Distill-Qwen-7B.
    }
  \label{tab:compatibility_main}
  \begin{tabular*}{\columnwidth}{@{\extracolsep{\fill}}lccc}
    \toprule
    \textbf{Method} 
    & \textbf{Accuracy (\%)} 
    & \textbf{\makecell{Training\\Time (s)}} 
    & \textbf{\makecell{Speedup\\(times)}} \\
    \midrule
    DAPO 
      & 87.13 
      & 42062 
      & 1.00 \\
    DAPO+\bppo{} 
      & 86.74 
      & 20430 
      & 2.06 \\
    \midrule
    GSPO 
      & 92.72 
      & 15287 
      & 1.00 \\
    GSPO+\bppo{} 
      & 91.99 
      & 2669 
      & 5.73 \\
    \bottomrule
  \end{tabular*}
\end{table}


\begin{table}[H]
\centering
\footnotesize
\setlength{\tabcolsep}{4pt}
\renewcommand{\arraystretch}{1.08}
\caption{Efficiency contribution of \bppo{} components on GSM8K with Qwen2.5-1.5B-Instruct.}
\label{tab:efficiency_components}
\begin{tabular*}{\columnwidth}{@{\extracolsep{\fill}}lccc}
\toprule
\textbf{Method} & \textbf{Accuracy (\%)} & \textbf{\makecell{Training\\Time (s)}}     &     \textbf{\makecell{Speedup\\(times)}} \\
\midrule
\grpo{}              & 76.63 & 12409 & 1.00 \\
\grpo{}+\firstn{}    & 76.44 & 11398 & 1.09 \\
Pair w/o ACS.      & 76.93 & 3942  & 3.15 \\
Pair w/ ACS.       & 76.62 & 2270  & 5.47 \\
\bppo{}              & 75.93 & 2021  & 6.14 \\
\bottomrule
\end{tabular*}
\end{table}

\subsection{Ablation Study}
\label{sec:ablation}


\paragraph{Efficiency Contributions of BPPO.}
Table~\ref{tab:efficiency_components} decomposes the efficiency gain of \bppo{} on GSM8K with Qwen2.5-1.5B-Instruct and \(G=16\). 
Applying \firstn{} alone gives a \(1.09\times\) speedup, while restricting updates to a single correct--incorrect pair already increases the speedup to \(3.15\times\) with comparable accuracy. 
Adaptive completion scheduling further improves the speedup to \(5.47\times\) by maintaining device utilization, and adding prefix-focused updates yields the full \bppo{} variant with the largest speedup of \(6.14\times\). Appendix~\ref{app:timing_breakdown} provides a wall-clock timing breakdown on MATH without adaptive completion scheduling.

\paragraph{Retained Completion Types.}

\begin{figure}[t]
    \centering
    \includegraphics[width=\columnwidth]{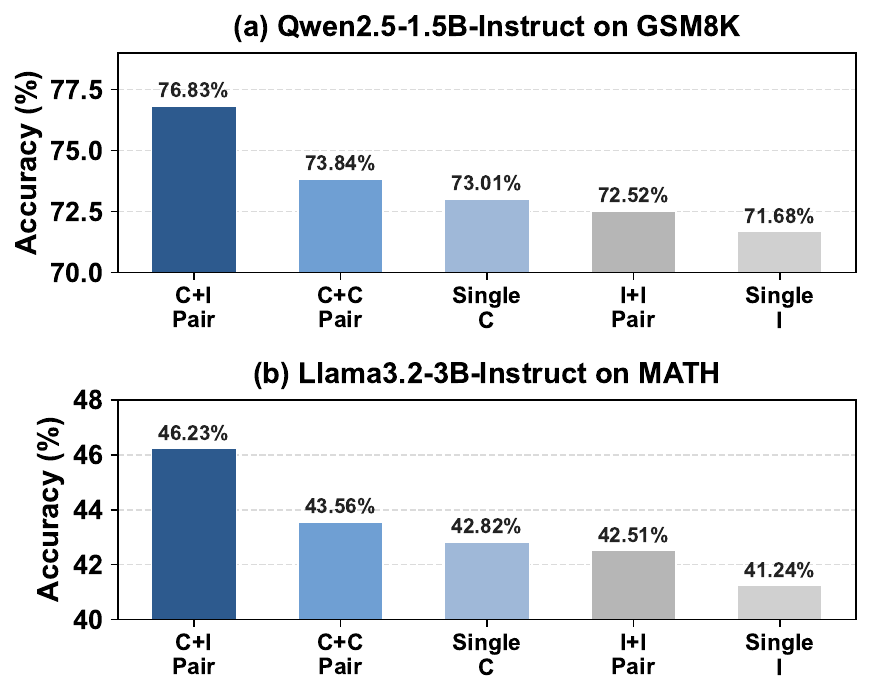}
    \caption{
    \textbf{Ablation on retained completion types.}
    (a) and (b) show results on GSM8K and MATH.
    The correct--incorrect (C+I) pair achieves the best accuracy, highlighting the importance of cross-class contrast.
    }
    
    \label{fig:pair_vs_single}
\end{figure}

In Figure~\ref{fig:pair_vs_single}, C and I denote correct and incorrect completions. All variants in this ablation randomly sample retained completions from the specified correctness classes. \textbf{C+I Pair} outperforms same-class variants that retain only correct or only incorrect completions on both GSM8K and MATH.

\paragraph{Pair-Selection Strategy.}

We compare five strategies: \textbf{Random}, \textbf{Shortest}, \textbf{Long-C + Short-I}, \textbf{Short-C + Long-I}, and \textbf{Longest}. 
As shown in Table~\ref{tab:pair_strategy}, \textbf{Random} gives the highest accuracy, while \textbf{Shortest} achieves comparable accuracy with shorter responses and lower training time. 
This is because \textbf{Shortest} provides a symmetric contrast: concise correct reasoning is reinforced, while concise but incorrect shortcuts are penalized, preventing collapse toward overly short responses. 
By contrast, \textbf{Short-C + Long-I} overemphasizes brevity and causes a large accuracy drop.

\begin{table}[t]
  \centering
  \footnotesize
  \setlength{\tabcolsep}{3pt}
  \renewcommand{\arraystretch}{1.08}
  \caption{Ablation of pair-selection strategies on the GSM8K.}
  \label{tab:pair_strategy}
  \begin{tabular}{l@{\hspace{6pt}}ccc}
    \toprule
    \textbf{Strategy} 
    & \textbf{\makecell{Accuracy\\(\%)}} 
    & \textbf{\makecell{Mean Response\\Tokens}} 
    & \textbf{\makecell{Training\\Time (s)}}  \\
    \midrule
    Random             & 76.83 & 248.31 & 2637 \\
    Shortest           & 76.62 & 159.15 & 2270 \\
    Long-C + Short-I   & 74.45 & 493.23 & 3630 \\
    Short-C + Long-I   & 39.83 & 116.87 & 1870 \\
    Longest            & 75.51 & 296.83 & 2847 \\
    \bottomrule
  \end{tabular}
\end{table}

\begin{figure}[!t]
    \centering
    \includegraphics[width=\columnwidth]{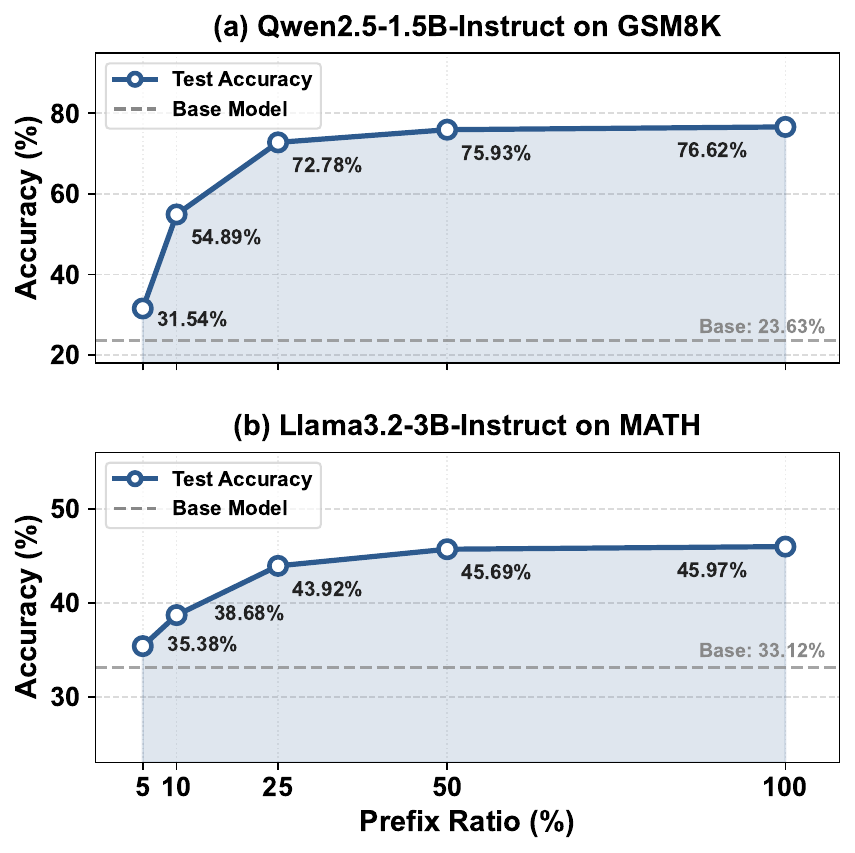}
    \caption{
    \textbf{Ablation on the retained prefix ratio in \firstn{}.}
    (a) and (b) show results on GSM8K and MATH, respectively.
    Accuracy improves with longer retained prefixes and largely saturates around 50\%, approaching full-response performance.
    }
    \label{fig:firstn_ratio}
\end{figure}

\paragraph{Prefix Length.}
The prefix length in \firstn{} controls how many response tokens contribute to the policy-gradient update. As shown in Figure~\ref{fig:firstn_ratio}, accuracy improves rapidly as the retained prefix ratio increases from 5\% to 25\%, and then gradually approaches full-response performance. At a 50\% prefix ratio, performance reaches 75.93\% on GSM8K and 45.69\% on MATH, close to the full-response results of 76.62\% and 45.97\%, respectively.

\section{Conclusion}
\label{sec:conclusion}

In this paper, we study how update signals are distributed within sampled groups in \grpo{}-style reasoning RL, where full-group and full-response updates are costly and may reinforce verbose trajectories. 
Our gradient-similarity analysis shows that, among completions sampled for the same prompt, same-class completions often induce similar update directions, while correct--incorrect pairs provide more distinct contrastive signals. 
Motivated by this, we propose \bppo{}, which retains the shortest correct--incorrect pair from each group as a compact update unit and applies policy updates only to the corresponding prefixes.
This reduces update-bearing completions and tokens while avoiding direct reinforcement of redundant reasoning patterns. 
Across GSM8K, MATH, and Geo3K, \bppo{} achieves up to \(5.90\times\), \(6.08\times\), and \(3.86\times\) training speedup over \grpo{}, respectively, while maintaining competitive accuracy and reducing mean response length by approximately 30--50\% without adding an explicit length penalty to the reward.

\section*{Limitations}
\bppo{} primarily accelerates training by reducing update-bearing completions and tokens, but it does not modify the generation procedure itself; thus, generation may still dominate the cost under long rollouts or expensive decoding. The prefix length is set to 50\% of the average response length as a simple heuristic, which may not be optimal across tasks, models, or training stages. \bppo{} also assumes that completions can be reliably divided into correct and incorrect classes; extending it to preference-based, partial-credit, or multi-dimensional rewards requires defining suitable contrastive pairs. Due to computational constraints, our experiments remain limited in scale, benchmarks, and modalities.

\bibliography{custom}

\clearpage
\appendix
\section{Appendix}

\begin{algorithm}[H]
\caption{\bppo{} Training}
\label{alg:bppo}
\begin{algorithmic}[1]
\State \textbf{Input:} dataset \(\mathcal{D}\), size \(N\), initial policy \(\pi_{\theta_0}\), group size \(G\), target budget \(C_{\mathrm{tar}}\), prefix length \(n\)
\State Policy model \(\pi_{\theta}\gets \pi_{\theta_0}\)
\State Reference model \(\pi_{\mathrm{ref}}\gets \pi_{\theta}\)
\State Set \(B_{\mathrm{sch}}\gets \left\lceil C_{\mathrm{tar}}/2 \right\rceil\), \(T\gets \left\lceil N/B_{\mathrm{sch}} \right\rceil\)

\For{step \(=1,\dots,T\)}
    \State Update old policy \(\pi_{\theta_{\mathrm{old}}}\gets \pi_{\theta}\)

    \Statex \textcolor{myblue}{\textbf{\# Adaptive completion scheduling}}
    \State Sample scheduled prompt batch \(\mathcal{D}_b\) of \(B_{\mathrm{sch}}\) prompts
    \State Initialize update batch \(\mathcal{B}\gets\emptyset\)

    \Statex \textcolor{myblue}{\textbf{\# Pair selection}}
    \For{each prompt \(q^{(j)}\in\mathcal{D}_b\)}
        \State Sample completions \(\{o_i^{(j)}\}_{i=1}^{G}\sim \pi_{\theta_{\mathrm{old}}}(\cdot\mid q^{(j)})\)
        \State Compute rewards \(\{r_i\}_{i=1}^{G}\) and advantages \(\{\hat{A}_i\}_{i=1}^{G}\)
        \State \(\mathcal{I}_j^+=\{i:r_i>0\}\), \(\mathcal{I}_j^-=\{i:r_i\le0\}\)
        \If{\(\mathcal{I}_j^+=\emptyset\) or \(\mathcal{I}_j^-=\emptyset\)}
            \State \textbf{continue}
        \EndIf
        \State \(i_{+}^{*}\gets\argmin_{i\in\mathcal{I}_j^+}|o_i^{(j)}|\)
        \State \(i_{-}^{*}\gets\argmin_{i\in\mathcal{I}_j^-}|o_i^{(j)}|\)
        \State \(\mathcal{P}_j\gets\{i_+^*,i_-^*\}\)
        \State Add selected pair \(\{(q^{(j)},o_i^{(j)},\hat{A}_i):i\in\mathcal{P}_j\}\) to \(\mathcal{B}\)
    \EndFor

    \Statex \textcolor{myblue}{\textbf{\# Prefix-focused update}}
    \State Keep the first \(n\) response tokens for each completion in \(\mathcal{B}\)
    \State Update \(\theta\) with the \bppo{} objective on \(\mathcal{B}\)
\EndFor

\State \textbf{Output:} \(\pi_{\theta}\)
\end{algorithmic}
\end{algorithm}

\setcounter{table}{7}
\begin{table*}[!t]
  \centering
  \small
  \setlength{\tabcolsep}{4.5pt}
  \renewcommand{\arraystretch}{1.06}
  \caption{Additional \(G=32\) results on GSM8K, MATH, and Geo3K. Each setting is trained three times independently, and we report mean \(\pm\) standard deviation. \bppo{} achieves the largest speedup and response-length reduction across all three benchmarks, with slight accuracy drops in some settings.}
  \label{tab:g32_results}
  \begin{tabular*}{0.98\textwidth}{@{\extracolsep{\fill}}llccccc}
    \toprule
    \textbf{\makecell{Benchmark\\Model}}
    & \textbf{Method}
    & \textbf{Accuracy (\%)}
    & \textbf{\makecell{Mean\\Response Tokens}}
    & \textbf{\makecell{Length\\Reduction (\%)}}
    & \textbf{\makecell{Training\\Time (s)}}
    & \textbf{\makecell{Speedup\\(times)}} \\
    \midrule

    \multirow{5}{*}{\makecell{GSM8K\\Qwen2.5-1.5B\\-Instruct}}
    & \grpo{}
    & \(76.96 \pm 0.35\)
    & \(265.08 \pm 2.58\)
    & -
    & \(23519 \pm 4033\)
    & 1.00 \\

    & CPPO
    & \(73.64 \pm 0.72\)
    & \(232.62 \pm 3.08\)
    & 12.2
    & \(3587 \pm 552\)
    & \(6.56 \pm 1.01\) \\

    & \grpo{}+\firstn{}
    & \(76.85 \pm 0.28\)
    & \(237.77 \pm 6.93\)
    & 10.3
    & \(20205 \pm 860\)
    & \(1.16 \pm 0.18\) \\

    & Pair
    & \(74.44 \pm 0.21\)
    & \(144.77 \pm 3.71\)
    & 45.4
    & \(3196 \pm 256\)
    & \(7.36 \pm 0.59\) \\

    & \bppo{}
    & \(73.84 \pm 1.05\)
    & \(122.39 \pm 5.46\)
    & \bestbox{53.8}
    & \(2672 \pm 51\)
    & \bestbox{\(\mathbf{8.80} \pm 0.17\)} \\

    \midrule

    \multirow{5}{*}{\makecell{MATH\\Llama3.2-3B\\-Instruct}}
    & \grpo{}
    & \(46.07 \pm 0.43\)
    & \(403.94 \pm 5.93\)
    & -
    & \(17763 \pm 1776\)
    & 1.00 \\

    & CPPO
    & \(45.28 \pm 0.15\)
    & \(359.80 \pm 14.21\)
    & 10.9
    & \(2553 \pm 180\)
    & \(6.96 \pm 0.49\) \\

    & \grpo{}+\firstn{}
    & \(46.74 \pm 0.17\)
    & \(371.49 \pm 6.07\)
    & 8.0
    & \(15519 \pm 187\)
    & \(1.14 \pm 0.12\) \\

    & Pair
    & \(45.02 \pm 0.30\)
    & \(163.15 \pm 0.94\)
    & 59.6
    & \(2382 \pm 162\)
    & \(7.46 \pm 0.51\) \\

    & \bppo{}
    & \(44.76 \pm 0.18\)
    & \(154.10 \pm 2.75\)
    & \bestbox{61.9}
    & \(2041 \pm 26\)
    & \bestbox{\(\mathbf{8.70} \pm 0.11\)} \\

    \midrule

    \multirow{5}{*}{\makecell{Geo3K\\Qwen3-VL-4B}}
    & \grpo{}
    & \(49.34 \pm 0.25\)
    & \(3001.24 \pm 29.15\)
    & -
    & \(39447 \pm 177\)
    & 1.00 \\

    & CPPO
    & \(49.56 \pm 0.68\)
    & \(4005.67 \pm 42.29\)
    & -33.5
    & \(22162 \pm 253\)
    & \(1.78 \pm 0.02\) \\

    & \grpo{}+\firstn{}
    & \(48.89 \pm 0.19\)
    & \(3025.72 \pm 15.25\)
    & -0.8
    & \(36419 \pm 73\)
    & \(1.08 \pm 0.01\) \\

    & Pair
    & \(49.73 \pm 0.08\)
    & \(1618.41 \pm 45.07\)
    & 46.1
    & \(20403 \pm 541\)
    & \(1.93 \pm 0.05\) \\

    & \bppo{}
    & \(49.85 \pm 0.86\)
    & \(1438.18 \pm 50.32\)
    & \bestbox{52.1}
    & \(18952 \pm 376\)
    & \bestbox{\(\mathbf{2.08} \pm 0.04\)} \\

    \bottomrule
  \end{tabular*}
\end{table*}

\setcounter{table}{8}
\begin{table*}[!t]
  \centering
  \small
  \setlength{\tabcolsep}{4.2pt}
  \renewcommand{\arraystretch}{1.08}
  \caption{
  Results without adaptive completion scheduling under rollout group size \(G=32\).
    \bppo{} still achieves accuracy comparable \grpo{} while updating only the prefixes of two selected completions from each \(G=32\) group.
  }
  \label{tab:no_adaptive_scheduling_g32}
  \begin{tabular*}{0.98\textwidth}{@{\extracolsep{\fill}}llccccc}
    \toprule
    \textbf{\makecell{Benchmark\\Model}}
    & \textbf{Method}
    & \textbf{Accuracy (\%)}
    & \textbf{\makecell{Mean\\Response Tokens}}
    & \textbf{\makecell{Length\\Reduction (\%)}}
    & \textbf{\makecell{Training\\Time (s)}}
    & \textbf{\makecell{Speedup\\(times)}} \\
    \midrule

    \multirow{4}{*}{\makecell{GSM8K\\Qwen2.5-1.5B\\-Instruct}}
    & \grpo{}
    & 76.96
    & 265.08
    & -
    & 23519
    & 1.00 \\

    & \grpo{}+\firstn{}
    & 76.85
    & 237.77
    & 10.3
    & 20205
    & 1.16 \\

    & Pair w/o ACS
    & 77.86
    & 175.28
    & 33.9
    & 4506
    & \(5.22\) \\

    & \bppo{} w/o ACS
    & 77.48
    & 164.31
    & \bestbox{38.0}
    & 4201
    & \bestbox{\(\mathbf{5.60}\)} \\

    \midrule

    \multirow{4}{*}{\makecell{MATH\\Llama3.2-3B\\-Instruct}}
    & \grpo{}
    & 46.07
    & 403.94
    & -
    & 17763
    & 1.00 \\

    & \grpo{}+\firstn{}
    & 46.74
    & 371.49
    & 8.0
    & 15519
    & 1.14 \\

    & Pair w/o ACS
    & 48.62
    & 236.03
    & 41.6
    & 10948
    & \(1.62\) \\

    & \bppo{} w/o ACS
    & 48.16
    & 214.14
    & \bestbox{47.0}
    & 10092
    & \bestbox{\(\mathbf{1.76}\)} \\

    \midrule

    \multirow{4}{*}{\makecell{Geo3K\\Qwen3-VL-4B}}
    & \grpo{}
    & 49.34
    & 3001.24
    & -
    & 39447
    & 1.00 \\

    & \grpo{}+\firstn{}
    & 48.89
    & 3025.72
    & -0.8
    & 36419
    & 1.08 \\

    & Pair w/o ACS
    & 49.27
    & 1853.20
    & 38.3
    & 28394
    & \(1.39\) \\

    & \bppo{} w/o ACS
    & 50.16
    & 1548.00
    & \bestbox{48.4}
    & 25354
    & \bestbox{\(\mathbf{1.56}\)} \\

    \bottomrule
  \end{tabular*}
\end{table*}

\subsection{Algorithm}
\label{app:algorithm}

Algorithm~\ref{alg:bppo} summarizes the \bppo{} training procedure.
At each step, \bppo{} samples a scheduled prompt batch \(\mathcal{D}_b\) so that the retained update-bearing completions match \(C_{\mathrm{tar}}\).
For each prompt, it samples \(G\) completions, computes full-group rewards and advantages, retains the shortest correct--incorrect pair, and discards all-correct or all-incorrect groups before importance-ratio computation.
The retained pairs form \(\mathcal{B}\), aggregating completions from more prompts and reducing optimization steps per epoch.
Policy updates use only the first \(n\) response tokens, while full completions are used for reward and advantage computation.

\section{Experiments}
\subsection{Additional Results on a Larger Model}
\label{app:larger_model}
As an additional scalability check, we evaluate \bppo{} on a larger model, Qwen3-14B, and report the results in Table~\ref{tab:qwen3_14b_gsm8k}. The results show that \bppo{} also remains effective at a larger scale, achieving a 5.22$\times$ speedup over \grpo{} while maintaining competitive accuracy.

\setcounter{table}{6}
\begin{table}[H]
\centering
\footnotesize
\setlength{\tabcolsep}{4pt}
\renewcommand{\arraystretch}{1.08}
\caption{
Results on GSM8K with Qwen3-14B.
\bppo{} achieves comparable accuracy to \grpo{} with substantially lower training time.
}
\label{tab:qwen3_14b_gsm8k}
\begin{tabular*}{\columnwidth}{@{\extracolsep{\fill}}lccc}
\toprule
\textbf{Method} & \textbf{Accuracy (\%)} & \textbf{\makecell{Training\\Time (s)}}     & \textbf{\makecell{Speedup\\(times)}} \\
\midrule
Qwen3-14B & 46.55 & -- & -- \\
\midrule
\grpo{} & 93.33 & 36598 & 1.00 \\
\bppo{} & 93.86 & 7006 & 5.22 \\
\bottomrule
\end{tabular*}
\end{table}

\begin{figure}[H]
    \centering
    \includegraphics[width=0.9\linewidth]{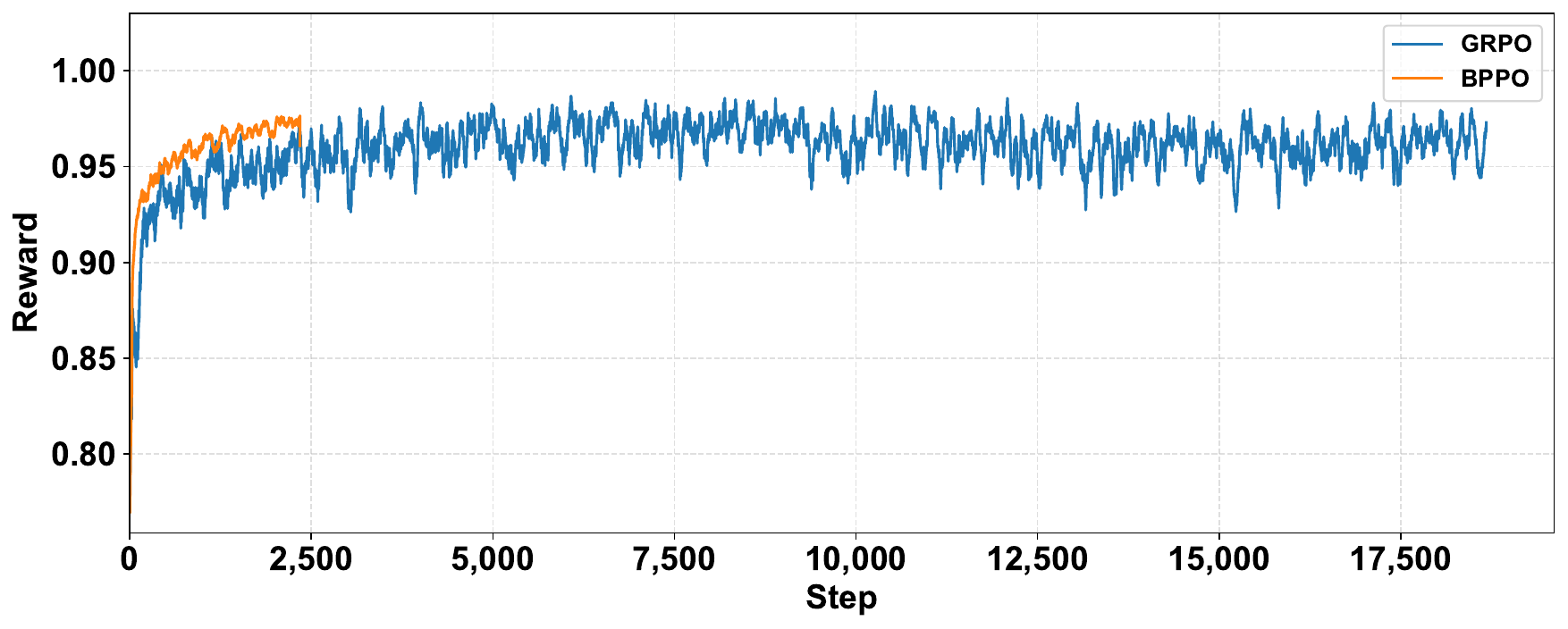}
    \caption{Reward curves of \grpo{} and \bppo{} on GSM8K with Qwen3-14B over 10 training epochs.}
    \label{fig:qwen3_14b_gsm8k_reward_curve}
\end{figure}

\begin{figure}[H]
    \centering
    \includegraphics[width=0.9\linewidth]{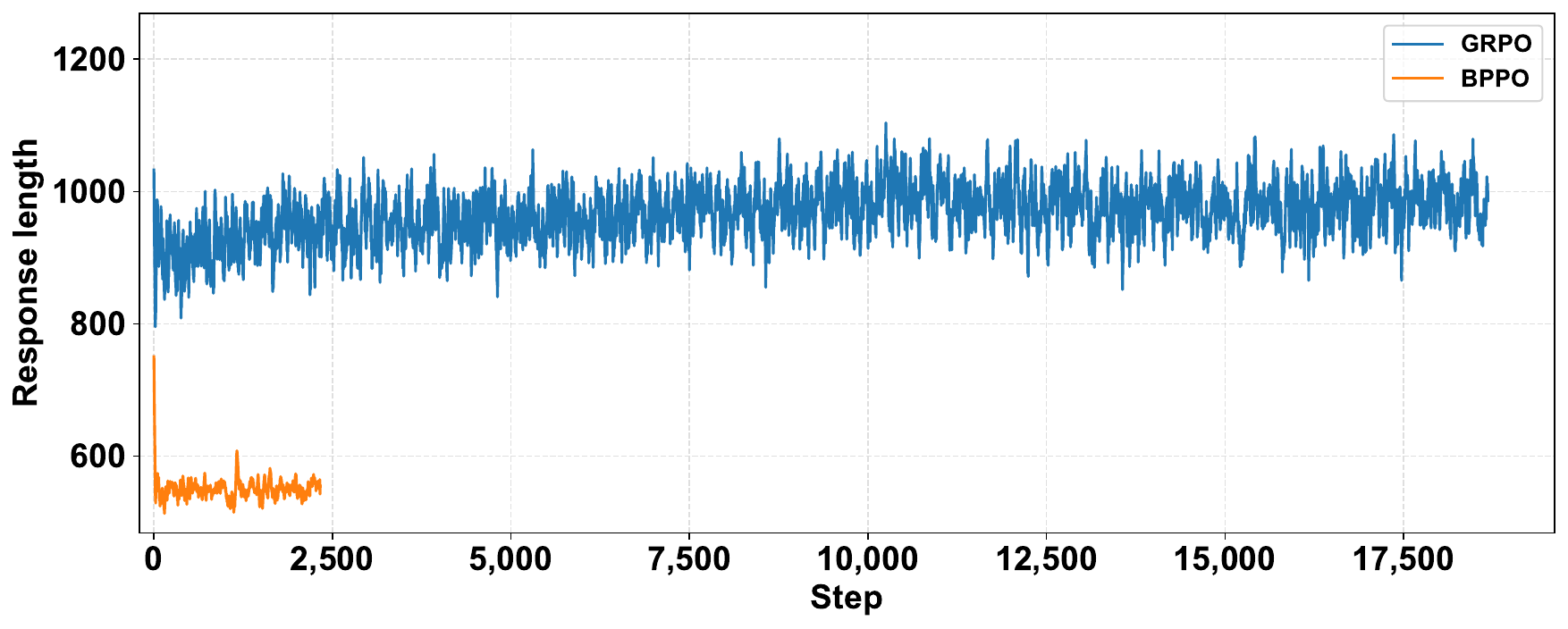}
    \caption{Response-length curves of \grpo{} and \bppo{} on GSM8K with Qwen3-14B over 10 training epochs.}
    \label{fig:qwen3_14b_gsm8k_length_curve}
\end{figure}

\paragraph{Optimization Stability and Convergence.}
To assess stability at a larger scale, we train both \grpo{} and \bppo{} for 10 epochs and plot their reward and response-length curves in Figures~\ref{fig:qwen3_14b_gsm8k_reward_curve} and~\ref{fig:qwen3_14b_gsm8k_length_curve}. \bppo{} shows convergence behavior comparable to \grpo{} without signs of instability or reward collapse. Its response length also remains stable and substantially lower than that of \grpo{} throughout training, indicating that \bppo{} maintains stable optimization behavior during large-model training.

\raggedbottom
\subsection{Additional Results for Larger Rollout Groups}
\label{app:g32_results}

We further report results under a larger rollout group size \(G=32\) in Table~\ref{tab:g32_results}. Larger rollout groups provide more candidates for shortest-pair selection, allowing \bppo{} to produce shorter responses. On text-only benchmarks, this also leads to stronger training speedups. \bppo{} consistently achieves the shortest responses, but also shows minor accuracy drops in some settings.

\subsection{Effect of Adaptive Completion Scheduling}
\label{app:acs_effect}

Adaptive completion scheduling reduces the number of optimization steps needed to process one training epoch. To isolate the effect of this scheduling component in a controlled comparison, we directly evaluate Pair and \bppo{} without adaptive completion scheduling, keeping the number of optimization steps matched with \grpo{} for the same training epoch under identical training settings. As shown in Table~\ref{tab:no_adaptive_scheduling_g32}, Pair and \bppo{} remain comparable to full-group \grpo{} in accuracy, while reducing response length and training time. These results show that, at \(G=32\), updating only the prefixes of the shortest correct--incorrect pair can still achieve accuracy comparable to full-group \grpo{} updates. Table~\ref{tab:no_adaptive_scheduling_g8_g16} further reports the corresponding matched-step results for \(G=8\) and \(G=16\).

\setcounter{table}{9}
\begin{table*}[!t]
  \centering
  \small
  \setlength{\tabcolsep}{4.2pt}
  \renewcommand{\arraystretch}{1.08}
  \caption{
  Results without adaptive completion scheduling under rollout group sizes \(G=8\) and \(G=16\).
  Pair and \bppo{} reduce response length and training time.
  }
  \label{tab:no_adaptive_scheduling_g8_g16}
  \begin{tabular*}{0.98\textwidth}{@{\extracolsep{\fill}}llcccccc}
    \toprule
    \textbf{Method}
    & \textbf{\makecell{Benchmark\\Model}}
    & \textbf{\makecell{Group\\Size \(G\)}}
    & \textbf{Accuracy (\%)}
    & \textbf{\makecell{Mean\\Response Tokens}}
    & \textbf{\makecell{Length\\Reduction (\%)}}
    & \textbf{\makecell{Training\\Time (s)}}
    & \textbf{\makecell{Speedup\\(times)}} \\
    \midrule

    Pair w/o ACS
    & \multirow{4}{*}{\makecell{GSM8K\\Qwen2.5-1.5B\\-Instruct}}
    & \multirow{2}{*}{8}
    & 76.82
    & 180.25
    & 30.4
    & 3388
    & \(2.12\) \\

    \bppo{} w/o ACS
    &
    &
    & 76.50
    & 129.61
    & \bestbox{50.0}
    & 3147
    & \bestbox{\(\mathbf{2.28}\)} \\

    Pair w/o ACS
    &
    & \multirow{2}{*}{16}
    & 77.33
    & 192.07
    & 26.9
    & 3942
    & \(3.00\) \\

    \bppo{} w/o ACS
    &
    &
    & 76.12
    & 125.84
    & \bestbox{52.1}
    & 3602
    & \bestbox{\(\mathbf{3.28}\)} \\

    \midrule

    Pair w/o ACS
    & \multirow{4}{*}{\makecell{MATH\\Llama3.2-3B\\-Instruct}}
    & \multirow{2}{*}{8}
    & 46.08
    & 253.84
    & 29.6
    & 8132
    & \(1.12\) \\

    \bppo{} w/o ACS
    &
    &
    & 45.84
    & 234.85
    & \bestbox{34.9}
    & 7820
    & \bestbox{\(\mathbf{1.16}\)} \\

    Pair w/o ACS
    &
    & \multirow{2}{*}{16}
    & 47.78
    & 231.00
    & 38.4
    & 8692
    & \(1.45\) \\

    \bppo{} w/o ACS
    &
    &
    & 46.20
    & 224.79
    & \bestbox{40.0}
    & 8382
    & \bestbox{\(\mathbf{1.51}\)} \\

    \midrule

    Pair w/o ACS
    & \multirow{4}{*}{\makecell{Geo3K\\Qwen3-VL-4B}}
    & \multirow{2}{*}{8}
    & 49.72
    & 2333.00
    & 23.5
    & 22152
    & \(1.13\) \\

    \bppo{} w/o ACS
    &
    &
    & 48.22
    & 2119.56
    & \bestbox{30.5}
    & 21085
    & \bestbox{\(\mathbf{1.19}\)} \\

    Pair w/o ACS
    &
    & \multirow{2}{*}{16}
    & 48.51
    & 1916.98
    & 37.5
    & 22809
    & \(1.28\) \\

    \bppo{} w/o ACS
    &
    &
    & 51.94
    & 1822.16
    & \bestbox{40.6}
    & 21665
    & \bestbox{\(\mathbf{1.35}\)} \\

    \bottomrule
  \end{tabular*}
\end{table*}

\setcounter{table}{10}
\begin{table*}[!t]
  \centering
  \small
  \setlength{\tabcolsep}{4.5pt}
  \renewcommand{\arraystretch}{1.35}
  \caption{
  Compatibility of \bppo{} with DAPO and GSPO on GSM8K and MATH.
  \bppo{} provides consistent efficiency gains across both methods.
  }
  \label{tab:additional_compatibility}
  \begin{tabular*}{0.98\textwidth}{@{\extracolsep{\fill}}cclccc}
    \toprule
    \textbf{Model} 
    & \textbf{Benchmark} 
    & \textbf{Method} 
    & \textbf{Accuracy (\%)} 
    & \textbf{\makecell{Training\\Time (s)}} 
    & \textbf{\makecell{Speedup\\(times)}} \\
    \midrule

    \multirow{10}{*}{Qwen3-8B}
      & \multirow{5}{*}{GSM8K} 
      & Baseline        & 32.90 & --    & -- \\[2pt]
      \cline{3-6}
      \addlinespace[2pt]
      &                    & DAPO            & 95.15 & 12991 & 1.00 \\
      &                    & DAPO+\firstn{}  & 94.16 & 11573 & 1.12 \\
      &                    & DAPO+Pair       & 95.38 & 5187  & 2.51 \\
      &                    & DAPO+\bppo{}    & 95.07 & 4441  & \bestbox{\(\mathbf{2.92}\)} \\
    \cline{2-6}
    \addlinespace[2pt]
      & \multirow{5}{*}{MATH} 
      & Baseline        & 60.64 & --    & -- \\[2pt]
      \cline{3-6}
      \addlinespace[2pt]
      &                    & DAPO            & 87.13 & 42062 & 1.00 \\
      &                    & DAPO+\firstn{}  & 87.04 & 38506 & 1.09 \\
      &                    & DAPO+Pair       & 86.86 & 22225 & 1.89 \\
      &                    & DAPO+\bppo{}    & 86.74 & 20430 & \bestbox{\(\mathbf{2.06}\)} \\

    \midrule

    \multirow{10}{*}{\makecell{DeepSeek-R1-Distill\\Qwen-7B}}
      & \multirow{5}{*}{GSM8K}
      & Baseline        & 27.41 & --    & -- \\[2pt]
      \cline{3-6}
      \addlinespace[2pt]
      &                    & GSPO            & 92.72 & 15287 & 1.00 \\
      &                    & GSPO+\firstn{}  & 93.18 & 13463 & 1.14 \\
      &                    & GSPO+Pair       & 92.34 & 3093  & 4.94 \\
      &                    & GSPO+\bppo{}    & 91.99 & 2669  & \bestbox{\(\mathbf{5.73}\)} \\
    \cline{2-6}
    \addlinespace[2pt]
      & \multirow{5}{*}{MATH}
      & Baseline        & 75.38 & --    & -- \\[2pt]
      \cline{3-6}
      \addlinespace[2pt]
      &                    & GSPO            & 83.08 & 37738 & 1.00 \\
      &                    & GSPO+\firstn{}  & 82.14 & 34333 & 1.10 \\
      &                    & GSPO+Pair       & 82.64 & 13475 & 2.80 \\
      &                    & GSPO+\bppo{}    & 81.18 & 12581 & \bestbox{\(\mathbf{3.00}\)} \\
    \bottomrule
  \end{tabular*}
\end{table*}

\subsection{Additional Compatibility Results}
\label{app:compatibility}

Table~\ref{tab:additional_compatibility} provides the complete compatibility results for DAPO and GSPO. Together with the representative results in Table~\ref{tab:compatibility_main}, these results show that the proposed pair-selection and prefix-focused update strategy is not tied to a specific policy objective. Instead, it can be combined with different reasoning-RL methods, consistently reducing training time across multiple reasoning benchmarks while largely preserving the accuracy of the original methods.

\setcounter{table}{11}
\begin{table*}[!t]
  \centering
  \small
  \setlength{\tabcolsep}{10pt}
  \renewcommand{\arraystretch}{1.08}
  \caption{
  Evaluation on additional mathematical reasoning benchmarks using Llama3.2-3B-Instruct after training on MATH for one epoch.
  We compare \grpo{} and \bppo{} under rollout group sizes \(G=8,16,32\).
\bppo{} achieves slightly higher average performance across the evaluated group sizes.
  }
  \label{tab:additional_math_benchmarks}
  \begin{tabular}{lccccc}
    \toprule
    \textbf{Method}
    & \textbf{\makecell{Group Size \(G\)}}
    & \textbf{AIME24}
    & \textbf{MinervaMATH}
    & \textbf{OlympiadBench}
    & \textbf{Average} \\
    \midrule

    \grpo{}
    & \multirow{2}{*}{8}
    & 13.33
    & 13.24
    & 10.83
    & 12.47 \\

    \bppo{}
    &
    & 13.33
    & 13.24
    & 11.42
    & \bestbox{12.66} \\

    \midrule

    \grpo{}
    & \multirow{2}{*}{16}
    & 6.67
    & 14.71
    & 10.36
    & 10.58 \\

    \bppo{}
    &
    & 10.00
    & 13.24
    & 11.57
    & \bestbox{11.60} \\

    \midrule

    \grpo{}
    & \multirow{2}{*}{32}
    & 3.33
    & 12.50
    & 11.42
    & 9.08 \\

    \bppo{}
    &
    & 6.67
    & 12.13
    & 11.72
    & \bestbox{10.17} \\

    \bottomrule
  \end{tabular}
\end{table*}

\begin{figure*}[!t]
\centering
\footnotesize
\setlength{\tabcolsep}{7pt}
\renewcommand{\arraystretch}{0.88}

\newcommand{\timingrow}[3]{
\noindent
\begin{minipage}[c]{0.52\linewidth}
    \centering
    \includegraphics[width=\linewidth]{#1}
\end{minipage}
\hspace{-0.2em}
\begin{minipage}[c]{0.39\linewidth}
    \raggedleft
    #2
\end{minipage}
#3
}

\timingrow
{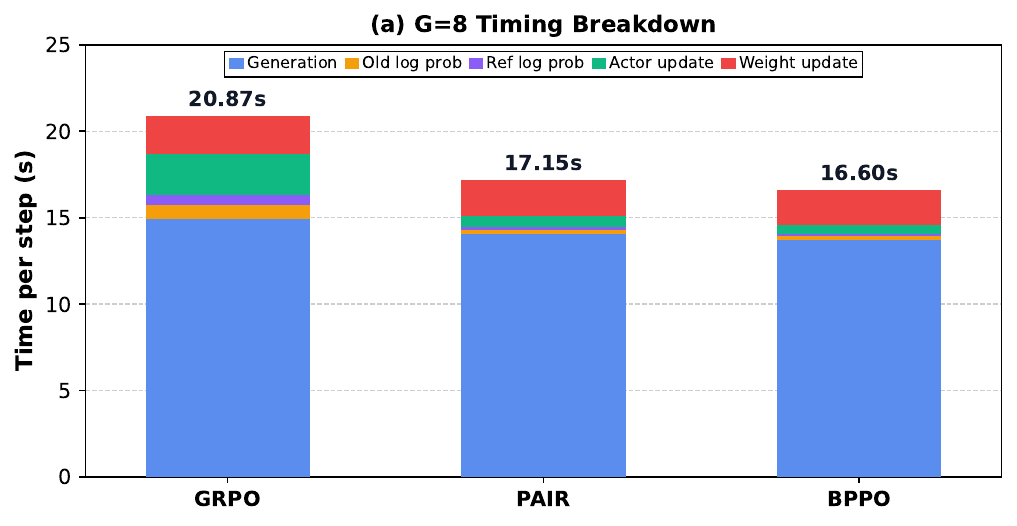}
{
\begin{tabular}{lccc}
\toprule
\multicolumn{4}{c}{$G=8$} \\
\midrule
Stage & GRPO & PAIR & BPPO \\
\midrule
Generation     & 14.95 & 14.03 & 13.72 \\
Old log prob   & 0.79  & 0.25  & 0.21  \\
Ref log prob   & 0.59  & 0.17  & 0.13  \\
Actor update   & 2.38  & 0.64  & 0.50  \\
Weight update  & 2.17  & 2.06  & 2.04  \\
\midrule
Total          & 20.87 & 17.15 & 16.60 \\
\bottomrule
\end{tabular}
}
{\\[-1.1em]}

\timingrow
{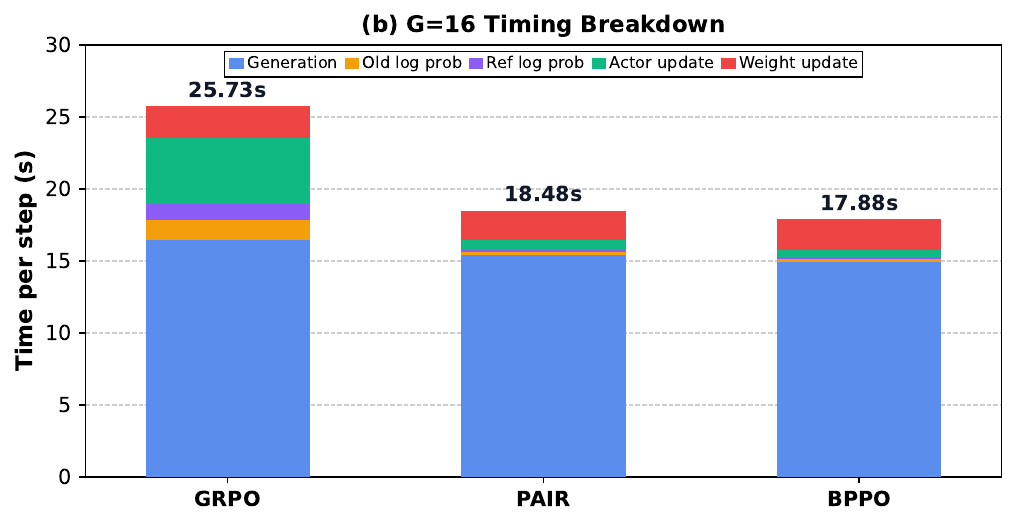}
{
\begin{tabular}{lccc}
\toprule
\multicolumn{4}{c}{$G=16$} \\
\midrule
Stage & GRPO & PAIR & BPPO \\
\midrule
Generation     & 16.47 & 15.38 & 14.94 \\
Old log prob   & 1.38  & 0.24  & 0.19  \\
Ref log prob   & 1.12  & 0.17  & 0.13  \\
Actor update   & 4.55  & 0.63  & 0.49  \\
Weight update  & 2.21  & 2.06  & 2.13  \\
\midrule
Total          & 25.73 & 18.48 & 17.88 \\
\bottomrule
\end{tabular}
}
{\\[-1.1em]}

\timingrow
{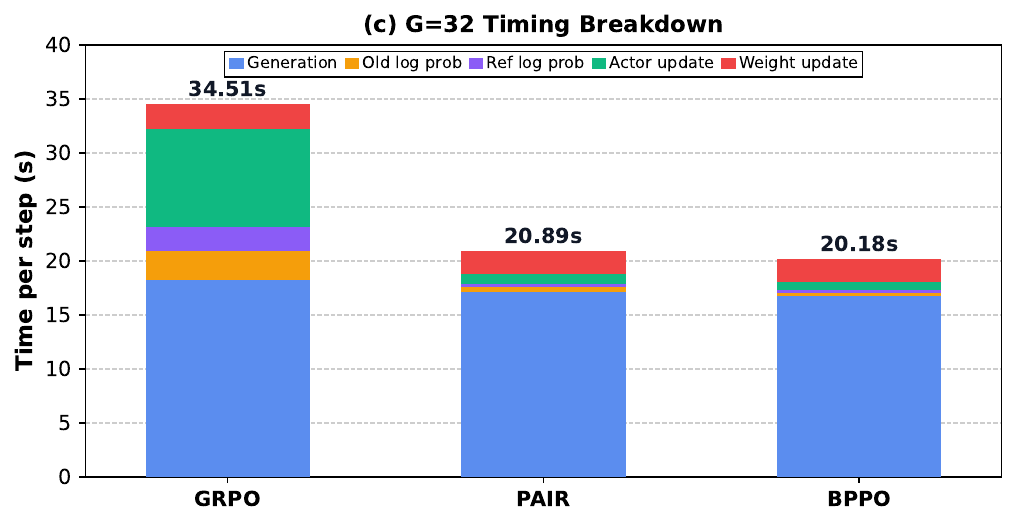}
{
\begin{tabular}{lccc}
\toprule
\multicolumn{4}{c}{$G=32$} \\
\midrule
Stage & GRPO & PAIR & BPPO \\
\midrule
Generation     & 18.19 & 17.12 & 16.72 \\
Old log prob   & 2.69  & 0.44  & 0.34  \\
Ref log prob   & 2.26  & 0.30  & 0.23  \\
Actor update   & 9.11  & 0.95  & 0.71  \\
Weight update  & 2.26  & 2.09  & 2.18  \\
\midrule
Total          & 34.51 & 20.89 & 20.18 \\
\bottomrule
\end{tabular}
}
{}

\vspace{-0.5em}
\caption{
Per-step wall-clock timing breakdown under different rollout group sizes.
}
\label{fig:appendix_timing_breakdown}
\end{figure*}

\raggedbottom

\subsection{Results on Additional Math Benchmarks}
\label{app:additional_math_benchmarks}

We further evaluate Llama3.2-3B-Instruct models trained on MATH for one epoch on additional mathematical reasoning benchmarks, including AIME24, MinervaMATH, and OlympiadBench. As shown in Table~\ref{tab:additional_math_benchmarks}, \bppo{} remains comparable to \grpo{} and achieves slightly higher performance across the evaluated rollout group sizes. These evaluations further support that shortest-pair prefix updates can match full-group \grpo{} performance.

\subsection{Wall-clock Timing Breakdown}
\label{app:timing_breakdown}

We report the per-step wall-clock timing breakdown of Llama-3.2-3B-Instruct on MATH under different rollout group sizes, as shown in Figure~\ref{fig:appendix_timing_breakdown}. Adaptive Completion Scheduling is disabled in this analysis to isolate the effect of pair selection and prefix truncation. The breakdown shows that \bppo{} substantially reduces the actor-update time across all rollout group sizes, which is the main source of per-step efficiency gains.

\onecolumn
\clearpage
\section{Case Study}
\label{app:case_study}
\begin{figure}[H]
\centering
\begin{minipage}{0.96\textwidth}

\begin{questionboxwide}
\textbf{Question.}
Find \(x\).

\vspace{0.4em}
\begin{center}
    \includegraphics[width=0.18\textwidth]{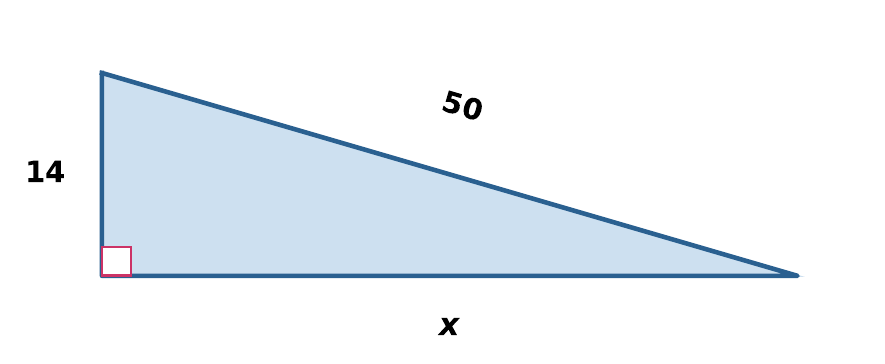}
\end{center}

\vspace{0.3em}
\textbf{Ground Truth.} \(48\)
\end{questionboxwide}

\vspace{0.7em}

\noindent
\begin{minipage}[t]{0.48\linewidth}
\vspace{0pt}
\begin{grpowidebox}
\footnotesize
\setlength{\abovedisplayskip}{3pt}
\setlength{\belowdisplayskip}{3pt}
I need to find \(x\) in this right triangle. The triangle has a right angle, with one leg labeled 14, the hypotenuse labeled 50, and the other leg labeled \(x\). This is a classic right triangle problem where I can use the Pythagorean theorem.

The Pythagorean theorem states that in a right triangle, the square of the hypotenuse, the side opposite the right angle, is equal to the sum of the squares of the other two sides. So, if I let the hypotenuse be \(c\), and the legs be \(a\) and \(b\), then
\[
a^2+b^2=c^2.
\]

In this case, the hypotenuse is 50, one leg is 14, and the other leg is \(x\). So, I can write the equation:
\[
14^2+x^2=50^2.
\]

Now I'll calculate the squares:
\[
196+x^2=2500.
\]

Subtract 196 from both sides:
\[
x^2=2500-196,
\]
\[
x^2=2304.
\]

Now take the square root of both sides:
\[
x=\sqrt{2304}.
\]

I know that \(48^2=2304\), because \(50^2=2500\) and
\[
48^2=(50-2)^2=2500-200+4=2304.
\]

So, \(x=48\).

Therefore, the value of \(x\) is 48.

I should note that since we're dealing with a length, \(x\) must be positive, so we take the positive square root.

Final answer: 48.

\textbf{Final Answer}
\[
\boxed{48}.
\]
\end{grpowidebox}
\end{minipage}
\hfill
\begin{minipage}[t]{0.48\linewidth}
\vspace{0pt}
\begin{bppowidebox}
\footnotesize
\setlength{\abovedisplayskip}{3pt}
\setlength{\belowdisplayskip}{3pt}
I need to find \(x\) in this right triangle. I see a right angle marked at the bottom left, with the side opposite the right angle labeled 14, and the hypotenuse labeled 50. The side adjacent to the right angle is labeled \(x\).

This is a right triangle, so I can use the Pythagorean theorem:
\[
a^2+b^2=c^2,
\]
where \(c\) is the hypotenuse.

Here, the hypotenuse is 50, one leg is 14, and the other leg is \(x\). So:
\[
14^2+x^2=50^2,
\]
\[
196+x^2=2500,
\]
\[
x^2=2500-196,
\]
\[
x^2=2304,
\]
\[
x=\sqrt{2304},
\]
\[
x=48.
\]

I can verify:
\[
14^2=196,\quad 48^2=2304,\quad 196+2304=2500.
\]
Perfect.

\textbf{Final Answer}
\[
\boxed{48}.
\]
\end{bppowidebox}
\end{minipage}

\vspace{0.7em}

\vspace{0.4em}

\caption{
Case study on Geo3K.
Both \grpo{} and \bppo{} generate the correct answer on a multimodal geometry problem, while \bppo{} provides a more compact derivation.
}
\label{fig:case_study_geo3k}

\end{minipage}
\end{figure}

We provide case studies from GSM8K, MATH, and Geo3K to illustrate the qualitative differences between responses generated by \grpo{} and \bppo{}.
On GSM8K, as shown in Figure~\ref{fig:case_study_gsm8k}, \bppo{} directly computes the daily total and scales it to a week, avoiding unnecessary unit conversions and intermediate explanations.
On MATH, as shown in Figure~\ref{fig:case_study_math_lunch}, \bppo{} captures the independent-choice structure directly, whereas \grpo{} enumerates all binomial cases, illustrating that a shorter path to the right answer can be the smarter choice.
On Geo3K, as shown in Figure~\ref{fig:case_study_geo3k}, \bppo{} focuses on the essential Pythagorean relation and gives a more compact derivation.
Together, these examples show that \bppo{} preserves correctness while removing redundant reasoning steps.

\begin{figure}[H]
\centering
\begin{minipage}{0.96\textwidth}

\begin{questionboxwide}
\textbf{Question.}
John takes care of 10 dogs. Each dog takes 0.5 hours a day to walk and take care of their business.
How many hours a week does he spend taking care of dogs?

\vspace{0.3em}
\textbf{Ground Truth.} \(35\)
\end{questionboxwide}

\vspace{0.7em}

\noindent
\begin{minipage}[t]{0.48\linewidth}
\vspace{0pt}
\begin{grpowidebox}
\footnotesize
\setlength{\abovedisplayskip}{3pt}
\setlength{\belowdisplayskip}{3pt}
To determine how many hours John spends taking care of dogs each week, we can break down the problem into smaller steps:

1. Calculate the daily time spent on one dog:

- Each dog takes 0.5 hours per day.

2. Calculate the total daily time for all dogs:

- John takes care of 10 dogs.

- Total daily time = Number of dogs \(\times\) Time per dog

- Total daily time = \(10 \times 0.5\) hours

3. Convert the daily time from hours to minutes (optional but useful):

- There are 60 minutes in an hour.

- Convert 0.5 hours to minutes:
\[
0.5 \text{ hours} \times 60 \text{ minutes/hour} = 30 \text{ minutes}
\]

- So, each dog takes 30 minutes per day.

4. Calculate the weekly time spent on all dogs:

- There are 7 days in a week.

- Total weekly time = Daily time \(\times\) Days per week

- Total weekly time =
(10 dogs) \(\times\) (30 minutes/dog/day) \(\times\) (7 days/week)

5. Perform the multiplication:

- First, multiply the number of dogs by the time per dog per day:
\[
10 \times 30 = 300 \text{ minutes}
\]

- Then, multiply this result by the number of days per week:
\[
300 \text{ minutes/day} \times 7 \text{ days/week}
= 2100 \text{ minutes/week}
\]

6. Convert the weekly time back to hours:

- There are 60 minutes in an hour.

- Convert 2100 minutes to hours:
\[
2100 \text{ minutes} \div 60 \text{ minutes/hour}
= 35 \text{ hours}
\]

Therefore, the final answer is:
\[
\boxed{35}.
\]
\end{grpowidebox}
\end{minipage}
\hfill
\begin{minipage}[t]{0.48\linewidth}
\vspace{0pt}
\begin{bppowidebox}
\footnotesize
\setlength{\abovedisplayskip}{3pt}
\setlength{\belowdisplayskip}{3pt}
To determine how many hours John spends taking care of dogs each week, we can break down the problem into smaller steps:

1. Time per dog per day = 0.5 hours.

2. Total time per day for 10 dogs:
\[
0.5 \times 10 = 5 \text{ hours}.
\]

3. Weekly time:
\[
5 \times 7 = 35 \text{ hours}.
\]

Therefore, the final answer is:
\[
\boxed{35}.
\]

John spends 35 hours a week taking care of his dogs.
\end{bppowidebox}
\end{minipage}

\vspace{0.7em}

\vspace{0.4em}

\caption{
Case study on GSM8K.
Both \grpo{} and \bppo{} generate the correct answer, while \bppo{} produces a more concise and direct reasoning trajectory.
}
\label{fig:case_study_gsm8k}

\end{minipage}
\end{figure}

\begin{figure}[H]
\centering
\begin{minipage}{0.96\textwidth}

\begin{questionboxwide}
\textbf{Question.}
A group of 10 Caltech students go to Lake Street for lunch.
Each student eats at either Chipotle or Panda Express.
In how many different ways can the students collectively go to lunch?

\vspace{0.3em}
\textbf{Ground Truth.} \(1024\)
\end{questionboxwide}

\vspace{0.7em}

\noindent
\begin{minipage}[t]{0.48\linewidth}
\vspace{0pt}
\begin{grpowidebox}
\footnotesize
\setlength{\abovedisplayskip}{3pt}
\setlength{\belowdisplayskip}{3pt}
To find the total number of ways the students can collectively go to lunch, we need to consider all possible combinations of students eating at Chipotle and Panda Express.

Let's use a variable to represent the number of students eating at Chipotle. We can let \(x\) be the number of students eating at Chipotle and \(10-x\) be the number of students eating at Panda Express.

Since each student eats at either Chipotle or Panda Express, the total number of ways the students can collectively go to lunch is the sum of the number of ways for each possible value of \(x\).

We can use the binomial coefficient formula to calculate the number of ways for each value of \(x\):

\[
\begin{aligned}
x=0 &: 1,\\
x=1 &: 10,\\
x=2 &: \binom{10}{2}=45,\\
x=3 &: \binom{10}{3}=120,\\
x=4 &: \binom{10}{4}=210,\\
x=5 &: \binom{10}{5}=252,\\
x=6 &: \binom{10}{6}=210,\\
x=7 &: \binom{10}{7}=120,\\
x=8 &: \binom{10}{8}=45,\\
x=9 &: \binom{10}{9}=10,\\
x=10 &: 1.
\end{aligned}
\]

Now, we can sum up the number of ways for each value of \(x\):

1+10+45+120+210+252+210+120+45+10+1=1024.

Therefore, the total number of ways the students can collectively go to lunch is 1024.

The final answer is
\[
\boxed{1024}.
\]
\end{grpowidebox}
\end{minipage}
\hfill
\begin{minipage}[t]{0.48\linewidth}
\vspace{0pt}
\begin{bppowidebox}
\footnotesize
\setlength{\abovedisplayskip}{3pt}
\setlength{\belowdisplayskip}{3pt}
To find the total number of ways the students can go to lunch, we can consider each student's choice as an independent event.

Each student has 2 choices: Chipotle or Panda Express. Since there are 10 students, the total number of ways the students can collectively go to lunch is given by the product of the number of choices for each student, which is \(2^{10}\).

Therefore, the total number of ways the students can go to lunch is
\[
2^{10}=\boxed{1024}.
\]

The final answer is:
\[
\boxed{1024}.
\]
\end{bppowidebox}
\end{minipage}

\vspace{0.7em}

\vspace{0.4em}

\caption{
Case study on MATH.
Both \grpo{} and \bppo{} generate the correct answer, while \bppo{} directly uses the fact that each student makes an independent binary choice, avoiding unnecessary enumeration.
}
\label{fig:case_study_math_lunch}

\end{minipage}
\end{figure}

\end{document}